\pdfoutput=1

\documentclass[11pt]{article}

\usepackage[]{ACL2023}

\usepackage{times}
\usepackage{latexsym}

\usepackage[T1]{fontenc}

\usepackage[utf8]{inputenc}

\usepackage{microtype}

\usepackage{inconsolata}
\usepackage{multicol}
\usepackage{multirow}
\usepackage{booktabs}
\usepackage{adjustbox}
\usepackage{hyperref}
\usepackage{url}
\usepackage{enumitem}
\usepackage{pifont}
\usepackage{amsmath}
\usepackage{amssymb}
\usepackage{amsfonts}
\newcommand{\cmark}{\ding{51}}%
\newcommand{\xmark}{\ding{55}}%

\title{KALM: Knowledge-Aware Integration of Local, Document, and Global Contexts for Long Document Understanding}

\author{Shangbin Feng$^1$ \ \ \ Zhaoxuan Tan$^2$ \ \ \ Wenqian Zhang$^2$ \ \ \ Zhenyu Lei$^2$ \ \ \ Yulia Tsvetkov$^1$ \\
$^1$University of Washington \ \ \ $^2$Xi'an Jiaotong University \\
\small \texttt{\{shangbin, yuliats\}@cs.washington.edu} \ \ \ \texttt{\{tanzhaoxuan, 2194510944, fischer\}@stu.xjtu.edu.cn}
}

\begin{document}
\maketitle
\begin{abstract}
With the advent of pretrained language models (LMs), increasing research efforts have been focusing on infusing commonsense and domain-specific knowledge to prepare LMs for downstream tasks. These works attempt to leverage knowledge graphs, the \textit{de facto} standard of symbolic knowledge representation, along with pretrained LMs.
While existing approaches have leveraged external knowledge, it remains an open question how to jointly incorporate knowledge graphs representing varying contexts—from local (e.g., sentence), to  document-level, to global knowledge—to enable knowledge-rich exchange across these contexts.
Such rich contextualization can be especially beneficial for long document understanding tasks since standard pretrained LMs are typically bounded by the input sequence length. 
In light of these challenges, we propose \textbf{KALM}, a \textbf{K}nowledge-\textbf{A}ware \textbf{L}anguage \textbf{M}odel that jointly leverages knowledge in local, document-level, and global contexts for long document understanding. KALM first encodes long documents and knowledge graphs into the three knowledge-aware context representations. It then processes each context with context-specific layers, followed by a ``context fusion'' layer that facilitates knowledge exchange to derive an overarching document representation. 
Extensive experiments demonstrate that KALM achieves state-of-the-art performance on six long document understanding tasks and datasets. 
Further analyses reveal that the three knowledge-aware contexts are complementary and they all contribute to model performance, while the importance and information exchange patterns of different contexts vary with respect to different tasks and datasets. \footnote{Code and data are publicly available at \url{https://github.com/BunsenFeng/KALM}.}
\end{abstract}

\section{Introduction}
Large language models (LMs) have become the dominant paradigm in NLP research, while knowledge graphs (KGs) are the \textit{de facto} standard of symbolic knowledge representation. Recent advances in knowledge-aware NLP focus on combining the two paradigms \citep{Kepler, GreaseLM, he2021klmo}, infusing encyclopedic \citep{DBLP:journals/cacm/VrandecicK14, DBLP:conf/esws/TanonWS20}, commonsense \citep{DBLP:conf/aaai/SpeerCH17}, and domain-specific \citep{DBLP:journals/corr/abs-2108-03861, DBLP:conf/bionlp/ChangBACBT20} knowledge with LMs. Knowledge-grounded models achieved state-of-the-art performance in tasks including question answering \citep{DBLP:conf/naacl/SunSQZ22}, commonsense reasoning \citep{DBLP:conf/naacl/KimKKAHY22, DBLP:conf/aaai/LiuW0PY21}, and social text analysis \citep{DBLP:conf/naacl/ZhangFCLLL22, DBLP:conf/acl/HuYZZTSD020}.


Prior approaches to infusing LMs with knowledge typically focused on three hitherto orthogonal directions: incorporating knowledge related to local (e.g., sentence-level), document-level, or global context. 
\textbf{Local} context approaches argue that sentences mention entities, and the external knowledge of entities, such as textual descriptions \citep{balachandran2021investigating, Kepler} and metadata \citep{ostapenko-etal-2022-speaker}, help LMs realize they are more than tokens.
\textbf{Document-level} approaches argue that core idea entities are repeatedly mentioned throughout the document, while related concepts might be discussed in different paragraphs. These methods attempt to leverage entities and knowledge across paragraphs with document graphs \citep{DBLP:journals/corr/abs-2108-03861, DBLP:conf/naacl/ZhangFCLLL22, DBLP:conf/acl/HuYZZTSD020}.
\textbf{Global} context approaches argue that unmentioned yet connecting entities help connect the dots for knowledge-based reasoning, thus knowledge graph subgraphs are encoded with graph neural networks alongside textual content \citep{GreaseLM, DBLP:conf/naacl/YasunagaRBLL21}. However, despite their individual pros and cons, how to integrate the three document contexts in a knowledge-aware way remains an open problem.

\begin{figure*}[]
    \centering
    \includegraphics[width=1\linewidth]{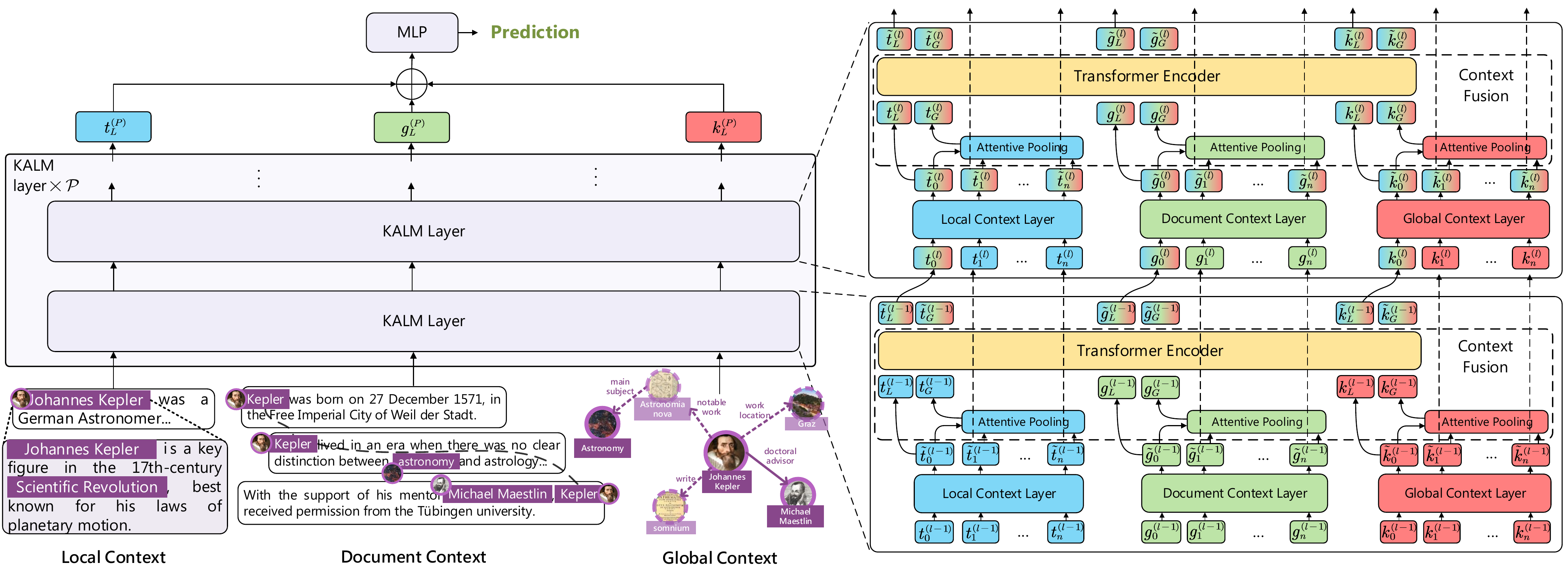}
    \caption{Overview of KALM, which encodes long documents and knowledge graphs into local, document, and global contexts while enabling information exchange across contexts.}
    \label{fig:overview}
\end{figure*}

Controlling for varying scopes of knowledge and context representations could benefit numerous language understanding tasks, especially those centered around long documents. 
Bounded by the inherent limitation of input sequence length, existing knowledge-aware LMs are mostly designed to handle short texts \citep{Kepler, GreaseLM}. However, processing long documents containing thousands of tokens \citep{beltagy2021beyond} requires attending to varying document contexts, disambiguating long-distance co-referring entities and events, and more.

In light of these challenges, we propose \textbf{KALM}, a \textbf{K}nowledge-\textbf{A}ware \textbf{L}anguage \textbf{M}odel for long document understanding. Specifically, KALM first derives  
three context- and knowledge-aware representations from the long input document and an external knowledge graph: 
the local context represented as raw text, the document-level context represented as a document graph, and the global context represented as a knowledge graph subgraph. KALM layers then encode each context with context-specific layers, followed by our proposed novel ContextFusion layers to enable knowledge-rich information exchange across the three knowledge-aware contexts. A unified document representation is then derived from context-specific representations that also interact with other contexts. An illustration of the proposed KALM is presented in Figure \ref{fig:overview}.

While KALM is a general method for long document understanding, we evaluate the model on six tasks and datasets that are particularly sensitive to broader contexts and external knowledge: political perspective detection, misinformation detection, and roll call vote prediction. Extensive experiments demonstrate that KALM outperforms pretrained LMs, task-agnostic knowledge-aware baselines, and strong task-specific baselines on all six datasets. In ablation experiments, we further establish KALM's ability to enable information exchange, better handle long documents, and improve data efficiency. In addition, KALM and the proposed ContextFusion layers reveal and help interpret the roles and information exchange patterns of different contexts.

\section{KALM Methodology}

\subsection{Problem Definition}
Let $\boldsymbol{d} = \{ \boldsymbol{d}_1, \ldots, \boldsymbol{d}_n \}$ denote a document with $n$ paragraphs, where each paragraph contains a sequence of $n_i$ tokens $\boldsymbol{d}_i = \{w_{i1}, \ldots, w_{in_{i}}\}$. Knowledge-aware long document understanding assumes the access to an external knowledge graph $KG = (\mathcal{E}, \mathcal{R}, \mathbf{A}, \epsilon, \varphi)$, where $\mathcal{E} = \{e_1, \ldots, e_{\mathcal{N}}\}$ denotes the entity set, $\mathcal{R} = \{r_1, \ldots, r_{\mathcal{M}}\}$ denotes the relation set, $\mathbf{A}$ is the adjacency matrix where $a_{ij} = k$ indicates $(e_i, r_k, e_j) \in KG$, $\epsilon(\cdot): \mathcal{E} \rightarrow \mathrm{str}$ and $\varphi(\cdot): \mathcal{R} \rightarrow \mathrm{str}$ map the entities and relations to their textual descriptions.

Given pre-defined document labels, knowledge-aware natural language understanding aims to learn document representations and classify $\boldsymbol{d}$ into its corresponding label with the help of $KG$.

\subsection{Knowledge-Aware Contexts}
We hypothesize that a holistic representation of long documents should incorporate contexts and relevant knowledge at three levels: the local context (e.g., a sentence with descriptions of mentioned entities), the broader document context (e.g., a long document with cross-paragraph entity reference structure), and the global/external context represented as external knowledge (e.g., relevant knowledge base subgraphs). Each of the three contexts uses different granularities of external knowledge, while existing works fall short of jointly integrating the three types of representations. To this end, KALM firstly employs different ways to introduce knowledge in different levels of contexts.

\paragraph{Local context.} Represented as the raw text of sentences and paragraphs, the local context models the smallest unit in long document understanding. 
Prior works attempted to add sentence metadata (e.g., tense, sentiment, topic) \citep{DBLP:conf/naacl/ZhangFCLLL22}, adopt sentence-level pretraining tasks based on KG triples \citep{Kepler}, or leverage knowledge graph embeddings along with textual representations \citep{DBLP:conf/acl/HuYZZTSD020}. While these methods were effective, in the face of LM-centered NLP research, they are ad-hoc add-ons and not fully compatible with existing pretrained LMs.
As a result, KALM proposes to directly concatenate the textual descriptions of entities $\epsilon(e_i)$ to the paragraph if $e_i$ is mentioned. In this way, the original text is directly augmented with the entity descriptions, informing the LM that entities such as "Kepler" are more than mere tokens and help to combat the spurious correlations of pretrained LMs \citep{DBLP:journals/corr/abs-2207-08982}. For each augmented paragraph $\boldsymbol{d}_i'$, we adopt $\mathrm{LM}(\cdot)$ and mean pooling to extract a paragraph representation. We use pretrained BART encoder \citep{DBLP:conf/acl/LewisLGGMLSZ20bart} as $\mathrm{LM}(\cdot)$ without further notice. We also add a fusion token at the beginning of the paragraph sequence for information exchange across contexts. After processing all $n$ paragraphs, we obtain the local context representation $\boldsymbol{T}^{(0)}$ as follows:

\begin{align*}
\begin{aligned}
    \boldsymbol{T}^{(0)} & = \{\mathbf{t}_0^{(0)}, \ldots, \mathbf{t}_n^{(0)}\} \\ 
    & = \{\mathbf{\theta}_{\textit{rand}}, \mathrm{LM}(\boldsymbol{d}_1'), \ldots, \mathrm{LM}(\boldsymbol{d}_n')\}
\end{aligned}
\end{align*}
where $\mathbf{\theta}_{\textit{rand}}$ denotes a randomly initialized vector of the fusion token in the local context and the superscript $(0)$ indicates the $0$-th layer.

\paragraph{Document-level context.} Represented as the structure of the full document, the document-level context is responsible for modeling cross-paragraph entities and knowledge on a document level. While existing works attempted to incorporate external knowledge in documents via document graphs \citep{DBLP:journals/corr/abs-2108-03861, DBLP:conf/acl/HuYZZTSD020}, they fall short of leveraging the overlapping entities and concepts between paragraphs that underpin the reasoning of long documents. To this end, we propose \emph{knowledge coreference}, a simple and effective mechanism for modeling text-knowledge interaction on the document level. Specifically, a document graph with $n+1$ nodes is constructed, consisting of one fusion node and $n$ paragraph nodes. If paragraph $i$ and $j$ both mention entity $e_k$ in the external KB, node $i$ and $j$ in the document graph are connected with relation type $k$. In addition, the fusion node is connected to every paragraph node with a super-relation. As a result, we obtain the adjacency matrix of the document graph $\mathbf{A}^g$. Paired with the knowledge-guided GNN to be introduced in Section \ref{subsec:kalmlayer}, knowledge coreference enables the information flow across paragraphs guided by external knowledge. Node feature initialization of the document graph is as follows:

\begin{align*}
\begin{aligned}
    \boldsymbol{G}^{(0)} & = \{\mathbf{g}_0^{(0)}, \ldots, \mathbf{g}_n^{(0)}\} \\
    & = \{\mathbf{\theta}_{\textit{rand}}, \mathrm{LM}(\boldsymbol{d}_1), \ldots, \mathrm{LM}(\boldsymbol{d}_n)\}
\end{aligned}
\end{align*}

\paragraph{Global context.} Represented as external knowledge graphs, the global context is responsible for leveraging unseen entities and facilitating KG-based reasoning. Existing works mainly focused on extracting knowledge graph subgraphs \citep{DBLP:conf/naacl/YasunagaRBLL21, GreaseLM} and encoding them alongside document content. Though many tricks are proposed to extract and prune KG subgraphs, in KALM, we employ a straightforward approach: for all mentioned entities in the long document, KALM merges their $k$-hop neighborhood to obtain a knowledge graph subgraph. We use $k=2$ following previous works \citep{GreaseLM, vashishth2019compositioncompgcn}, striking a balance between KB structure and computational efficiency while KALM could support any $k$ settings. A fusion entity is then introduced and connected with every other entity, resulting in a connected graph. In this way, KALM cuts back on the preprocessing for modeling global knowledge and better preserve the information in the KG. Knowledge graph embedding methods \citep{DBLP:conf/nips/BordesUGWY13transe} are then adopted to initialize node features of the KG subgraph:

\begin{align*}
\begin{aligned}
    \boldsymbol{K}^{(0)} & = \{\mathbf{k}_0^{(0)}, \ldots, \mathbf{k}_{|\rho(\boldsymbol{d})|}^{(0)}\} \\
    & = \{\mathbf{\theta}_{\textit{rand}}, \mathrm{KGE}(e_1), \ldots, \mathrm{KGE}(e_{|\rho(\boldsymbol{d})|})\}
\end{aligned}
\end{align*}
where $\mathrm{KGE}(\cdot)$ denotes the knowledge graph embeddings trained on the original KG, $|\rho(\boldsymbol{d})|$ indicates the number of mentioned entities identified in document $\boldsymbol{d}$. We use TransE \citep{DBLP:conf/nips/BordesUGWY13transe} to learn KB embeddings and use them for $\mathrm{KGE}(\cdot)$, while the knowledge base embeddings are kept frozen in the KALM training process.

\subsection{KALM Layers}
\label{subsec:kalmlayer}
After obtaining the local, document-level, and global context representations of long documents, we employ KALM layers to learn document representations. Specifically, each KALM layer consists of three context-specific layers to process each context. A ContextFusion layer is then adopted to enable the knowledge-rich information exchange across the three contexts.

\subsubsection{Context-Specific Layers}

\paragraph{Local context layer.} The local context is represented as a sequence of vectors extracted from the knowledge-enriched text with the help of pretrained LMs. We adopt transformer encoder layers \citep{DBLP:conf/nips/VaswaniSPUJGKP17} to encode the local context:

\begin{align*}
\begin{aligned}
    \Tilde{\boldsymbol{T}}^{(\ell)} & = \{\Tilde{\mathbf{t}}_0^{(\ell)}, \ldots, \Tilde{\mathbf{t}}_n^{(\ell)}\} \\
    & = \phi\Big(\mathrm{TrmEnc}(\{\mathbf{t}_0^{(\ell)}, \ldots, \mathbf{t}_n^{(\ell)}\})\Big)
\end{aligned}
\end{align*}
where $\phi(\cdot)$ denotes non-linearity, $\mathrm{TrmEnc}$ denotes the transformer encoder layer, and $\Tilde{\mathbf{t}}_0^{(\ell)}$ denotes the transformed representation of the fusion token. We omit the layer subscript $(\ell)$ for brevity.

\paragraph{Document-level context layer.} The document-level context is represented as a document graph based on knowledge coreference. To better exploit the entity-based relations in the document graph, we propose a knowledge-aware GNN architecture to enable \textbf{knowledge-guided message passing} on the document graph:

\begin{align*}
    \Tilde{\boldsymbol{G}} = \{\Tilde{\mathbf{g}}_0, \ldots, \Tilde{\mathbf{g}}_n = \mathrm{GNN}\Big(\{\mathbf{g}_0, \ldots, \mathbf{g}_n\}\Big)
\end{align*}
where $\mathrm{GNN}(\cdot)$ denotes the proposed knowledge-guided graph neural networks as follows:

\begin{align*}
    \Tilde{\mathbf{g}}_i = \phi\Big( \alpha_{i,i} \mathbf{\Theta} \mathbf{g}_i + \sum_{j \in \mathcal{N}(i)} \mathbf{\Theta} \mathbf{g}_j \Big)
\end{align*}
where $\alpha_{i,j}$ denotes the knowledge-guided attention weight and is defined as follows:

\begin{align*}
\resizebox{1\linewidth}{!}{
    $\alpha_{i,j} = \frac{\mathrm{exp}\Big( \mathrm{ELU}(\mathbf{a}^T [\mathbf{\Theta}\mathbf{g}_i || \mathbf{\Theta}\mathbf{g}_j || \mathbf{\Theta} f(\mathrm{KGE}(a_{ij}^g)) ] ) \Big)}{\sum_{k \in \mathcal{N}(i)} \mathrm{exp} \Big( \mathrm{ELU}( \mathbf{a}^T [\mathbf{\Theta}\mathbf{g}_i || \mathbf{\Theta}\mathbf{g}_k || \mathbf{\Theta} f(\mathrm{KGE}(a_{ik}^g))] )\Big)  }$
    }
\end{align*}
where $\Tilde{\mathbf{g}}_0$ denotes the transformed representation of the fusion node, $\mathbf{a}$ and $\mathbf{\Theta}$ are learnable parameters, $a_{ij}^g$ is the $i$-th row and $j$-th column value of adjacency matrix $\mathbf{A}^g$ of the document graph, $\mathrm{ELU}$ denotes the exponential linear unit activation function \citep{clevert2015fastelu}, and $f(\cdot)$ is a learnable linear layer. $\mathbf{\Theta}f(\mathrm{KGE}(a_{ij}^g))$ is responsible for enabling the knowledge-guided message passing on the document graph, enabling KALM to incorporate the entity and concept patterns in different paragraphs and their document-level interactions.

\paragraph{Global context layer.} The global context is represented as a relevant knowledge graph subgraph. We follow previous works and adopt GATs \citep{DBLP:conf/iclr/VelickovicCCRLB18} to encode the global context:

\begin{align*}
\begin{aligned}
    \Tilde{\boldsymbol{K}} & = \{\Tilde{\mathbf{k}}_0, \ldots, \Tilde{\mathbf{k}}_{|\rho(\boldsymbol{d})|}\} \\
    & = \mathrm{GAT}\Big(\{\mathbf{k}_0, \ldots, \mathbf{k}_{|\rho(\boldsymbol{d})|}\}\Big)
\end{aligned}
\end{align*}
where $\Tilde{\mathbf{k}}_0$ denotes the transformed representation of the fusion entity.

\subsubsection{ContextFusion Layer}
\label{subsec:contextfusion}
The local, document, and global contexts model external knowledge within sentences, across the document, and beyond the document. These contexts are closely connected and a robust long document understanding method should reflect their interactions.
Existing approaches mostly leverage only one or two of the contexts \citep{Kepler, DBLP:journals/corr/abs-2108-03861, DBLP:conf/naacl/ZhangFCLLL22}, falling short of jointly leveraging the three knowledge-aware contexts. In addition, they mostly adopted direct concatenation or MLP layers \citep{DBLP:conf/naacl/ZhangFCLLL22, GreaseLM, DBLP:conf/acl/HuYZZTSD020}, falling short of enabling context-specific information to flow across contexts in a knowledge-rich manner. As a result, we propose the ContextFusion layer to tackle these challenges. We firstly take a local perspective and extract the representations of the fusion tokens, nodes, and entities in each context:

\begin{align*}
    \Big[ \mathbf{t}_{L}, \mathbf{g}_{L}, \mathbf{k}_{L} \Big] = \Big[ \Tilde{\mathbf{t}}_0, \Tilde{\mathbf{g}}_0, \Tilde{\mathbf{k}}_0 \Big]
\end{align*}

We then take a global perspective and use the fusion token/node/entity as the query to conduct attentive pooling $\mathrm{ap}(\cdot,\cdot)$ across all other tokens/nodes/entities in each context:

\begin{align*}
\begin{aligned}
    \Big[ \mathbf{t}_G, \mathbf{g}_G, \mathbf{k}_G \Big] = \Big[ \mathrm{ap}\big(\Tilde{\mathbf{t}}_0, \{ \Tilde{\mathbf{t}}_i \}_{i=1}^n\big), \\ \mathrm{ap}\big(\Tilde{\mathbf{g}}_0, \{ \Tilde{\mathbf{g}}_i \}_{i=1}^n\big), \mathrm{ap}\big(\Tilde{\mathbf{k}}_0, \{ \Tilde{\mathbf{k}}_i \}_{i=1}^n\big) \Big]
\end{aligned}
\end{align*}
where attentive pooling $\mathrm{ap}(\cdot,\cdot)$ is defined as:

\begin{align*}
    \mathrm{ap}\big( \mathbf{q}, \{\mathbf{k}_i\}_{i=1}^n \big) = \sum_{i=1}^n \frac{\mathrm{exp}\big(\mathbf{q}\cdot\mathbf{k}_i\big)}{\sum_{j=1}^n \mathrm{exp}\big(\mathbf{q}\cdot\mathbf{k}_j\big)} k_i
\end{align*}

In this way, the fusion token/node/entity in each context serves as the information exchange portal. We then use a transformer encoder layer to enable information exchange across the contexts:

\begin{align*}
\begin{aligned}
    & \Big[\Tilde{\mathbf{t}}_{L}, \Tilde{\mathbf{g}}_{L}, \Tilde{\mathbf{k}}_{L}, \Tilde{\mathbf{t}}_G, \Tilde{\mathbf{g}}_G, \Tilde{\mathbf{k}}_G \Big] \\ & = \phi\Big(\mathrm{TrmEnc}\Big(\Big[\mathbf{t}_{L}, \mathbf{g}_{L}, \mathbf{k}_{L}, \mathbf{t}_G, \mathbf{g}_G, \mathbf{k}_G\Big]\Big)\Big)
\end{aligned}
\end{align*}

As a result, $\Tilde{\mathbf{t}}_{L}$, $\Tilde{\mathbf{g}}_{L}$, and $\Tilde{\mathbf{k}}_{L}$ are the representations of the fusion token/node/entity that incorporates information from other contexts. We formulate the output of the $l$-th layer as follows:

\begin{align*}
\begin{aligned}
    & \boldsymbol{T}^{(\ell+1)} = \{\Tilde{\mathbf{t}}_{L}^{(\ell)}, \Tilde{\mathbf{t}}_1^{(\ell)}, \ldots, \Tilde{\mathbf{t}}_n^{(\ell)} \}, \\
    & \boldsymbol{G}^{(\ell+1)} = \{\Tilde{\mathbf{g}}_{L}^{(\ell)}, \Tilde{\mathbf{g}}_1^{(\ell)}, \ldots, \Tilde{\mathbf{g}}_n^{(\ell)} \}, \\
    & \boldsymbol{K}^{(\ell+1)} = \{\Tilde{\mathbf{k}}_{L}^{(\ell)}, \Tilde{\mathbf{k}}_1^{(\ell)}, \ldots, \Tilde{\mathbf{k}}_n^{(\ell)} \}
\end{aligned}
\end{align*}

Our proposed ContextFusion layer is interactive since it enables the information to flow across different document contexts, instead of direct concatenation or hierarchical processing. The attention weights in $\mathrm{TrmEnc}(\cdot)$ of the ContextFusion layer could also provide insights into the roles and importance of each document context, which will be further explored in Section \ref{subsec:exchange}. To the best of our knowledge, KALM is the first work to jointly consider the three levels of document context and enable information exchange across document contexts.

\renewcommand{\arraystretch}{0.9}
\begin{table*}[t]
    \centering
    \caption{Model performance on three tasks and six datasets. Acc, MaF, miF, and BAcc denote accuracy, macro-averaged F1-score, micro-averaged F1-score, and balanced accuracy. Best performance is shown in \textbf{bold}. Certain task-specific models did not report standard deviation in the original paper.}
    \resizebox{1\linewidth}{!}{
    \begin{tabular}{lllccccccccc}
        \toprule[1.5pt]
        \multirow{2}{*}{\textbf{Task}} & \multirow{2}{*}{\textbf{Dataset}} & \multirow{2}{*}{\textbf{Metric}} & \multirow{2}{*}{\textbf{Task SOTA}} & \multirow{2}{*}{\textbf{Best LM}} & \multicolumn{6}{c}{\textbf{Knowledge-Aware LMs}} & \multirow{2}{*}{\textbf{KALM}} \\ 
        \cmidrule[0.75pt](lr){6-11}
        & & & & & KELM & KnowBERT & Joshi et al. & KGAP & GreaseLM & GreaseLM+ &  \\
        \midrule[0.75pt]
        \multirow{4}{*}{\textbf{PDD}} & \multirow{2}{*}{\textbf{SemEval}} & \textbf{Acc} & $89.90~(\pm 0.6)$& $86.99~(\pm 1.9)$& $86.40~(\pm 2.3)$& $84.73~(\pm 3.4)$& $81.88~(\pm 2.1)$ & $87.73~(\pm 1.8)$& $86.64~(\pm 1.5)$ & $85.66~(\pm 1.8)$ & $\textbf{91.45}~(\pm 0.8)$ \\
        & & \textbf{MaF} & $86.11~(\pm 1.1)$& $80.62~(\pm 3.8)$& $83.98~(\pm 1.0)$& $75.72~(\pm 5.3)$& $77.15~(\pm 3.8)$& $82.00~(\pm 3.1)$& $80.32~(\pm 3.0)$ & $77.23~(\pm 4.1)$ & $\textbf{87.65}~(\pm 1.2)$\\
        & \multirow{2}{*}{\textbf{Allsides}} & \textbf{Acc} & $87.17~(\pm 0.2)$& $68.71~(\pm 4.3)$& $80.71~(\pm 2.4)$& $60.56~(\pm 0.7)$& $80.88~(\pm 2.1)$& $83.65~(\pm 1.3)$& $80.23~(\pm 1.2)$ & $82.16~(\pm 5.5)$ & $\textbf{87.26}~(\pm 0.2)$\\
        & & \textbf{MaF} & $86.72~(\pm 0.3)$& $65.39~(\pm 5.7)$& $79.74~(\pm 2.7)$& $58.81~(\pm 0.5)$& $79.73~(\pm 2.3)$& $82.92~(\pm 1.4)$& $79.17~(\pm 1.2)$ & $80.81~(\pm 7.1)$ & $\textbf{86.79}~(\pm 0.2)$\\
        \midrule[0.75pt]
        \multirow{4}{*}{\textbf{MD}} & \multirow{2}{*}{\textbf{SLN}} & \textbf{MiF} & $89.17$& $88.17~(\pm 0.6)$& $84.11~(\pm 0.6)$& $78.67~(\pm 3.2)$& $82.72~(\pm 5.1)$ & $92.17~(\pm 1.2)$& $73.83~(\pm 0.9)$ & $88.17~(\pm 0.8)$ & $\textbf{94.22}~(\pm 1.2)$\\
        & & \textbf{MaF} & $89.12$& $88.46~(\pm 4.9)$& $82.80~(\pm 1.3)$& $79.80~(\pm 2.0)$& $83.98~(\pm 3.7)$& $92.30~(\pm 0.9)$& $75.20~(\pm 0.8)$ & $88.64~(\pm 0.6)$ & $\textbf{94.18}~(\pm 1.1)$\\
        & \multirow{2}{*}{\textbf{LUN}} & \textbf{MiF} & $69.05$& $60.10~(\pm 1.7)$& $59.28~(\pm 2.1)$& $59.66~(\pm 1.1)$ & $58.57~(\pm 3.4)$& $65.52~(\pm 2.3)$& $56.54~(\pm 1.5)$ & $64.29~(\pm 2.4)$ & $\textbf{71.28}~(\pm 1.7)$\\
        & & \textbf{MaF} & $68.26$& $58.57~(\pm 2.1)$& $57.30~(\pm 1.6)$& $59.19~(\pm 1.3)$& $56.73~(\pm 4.0)$& $63.94~(\pm 2.9)$& $55.75~(\pm 1.6)$ & $62.65~(\pm 3.7)$ & $\textbf{69.82}~(\pm 1.2)$\\
        \midrule[0.75pt]
        \multirow{4}{*}{\textbf{RCVP}} & \multirow{2}{*}{\textbf{Random}} & \textbf{BAcc} & $90.33$& $89.94~(\pm 0.2)$& $89.13~(\pm 1.1)$& $86.72~(\pm 0.9)$ & $92.43~(\pm 0.5)$& $77.98~(\pm 0.5)$& $89.99~(\pm 1.5)$ & $91.01~(\pm 0.2)$ & $\textbf{92.36}~(\pm 0.4)$\\
        & & \textbf{MaF} & $84.92$& $86.10~(\pm 0.7)$& $84.76~(\pm 2.0)$& $79.33~(\pm 2.4)$& $89.64~(\pm 0.6)$& $68.11~(\pm 6.0)$& $84.72~(\pm 3.0)$ & $87.29~(\pm 0.3)$ & $\textbf{89.33}~(\pm 0.4)$\\
        & \multirow{2}{*}{\textbf{Time-based}} & \textbf{BAcc} & $89.92$& $90.40~(\pm 0.8)$& $90.80~(\pm 0.2)$& $87.07~(\pm 0.9)$& $92.63~(\pm 1.6)$& $77.90~(\pm 0.6)$& $88.21~(\pm 2.7)$ & $91.69~(\pm 0.1)$ & $\textbf{94.46}~(\pm 0.4)$\\
        & & \textbf{MaF} & $84.35$& $85.21~(\pm 2.1)$& $86.62~(\pm 0.4)$& $78.90~(\pm 1.9)$& $89.31~(\pm 2.4)$& $70.81~(\pm 4.6)$& $79.73~(\pm 7.4)$ & $87.95~(\pm 0.3)$ & $\textbf{91.97}~(\pm 0.5)$\\
        \bottomrule[1.5pt]
    \end{tabular}
    \label{tab:big_table}
    }
\end{table*}

\renewcommand{\arraystretch}{0.9}
\begin{table*}[t]
    \centering
    \caption{Ablation study of the three document contexts and the ContextFusion layer. Best performance is shown in \textbf{bold}. The local, document, and global contexts all contribute to model performance, while the ContextFusion layer is better than existing strategies at enabling information exchange across contexts.}
    \resizebox{1\linewidth}{!}{
    \begin{tabular}{lllccccccc}
        \toprule[1.5pt]
        \multicolumn{1}{c}{\multirow{2}{*}{\textbf{Task}}} & \multicolumn{1}{c}{\multirow{2}{*} {\textbf{Dataset}}} & \multicolumn{1}{c}{\multirow{2}{*}{\textbf{Metric}}} & \multicolumn{1}{c}{\textbf{Ours}} & \multicolumn{3}{c}{\textbf{Remove Context}} & \multicolumn{3}{c}{\textbf{Substitute ContextFusion}} \\ 
        \cmidrule[0.75pt](lr){4-4}
        \cmidrule[0.75pt](lr){5-7}
        \cmidrule[0.75pt](lr){8-10}
        & & & \multicolumn{1}{c}{KALM} & w/o local & w/o document & \multicolumn{1}{c}{w/o global} & MInt & concat & sum \\
        \midrule[0.75pt]
        \multirow{4}{*}{\textbf{PDD}} & \multirow{2}{*}{\textbf{SemEval}} & \textbf{Acc} & $\textbf{91.45}~(\pm 0.8)$& $83.55~(\pm 0.8)$& $83.57~(\pm 1.1)$& $84.11~(\pm 0.9)$& $81.91~(\pm 0.9)$ & $83.52~(\pm 1.8)$& $83.21~(\pm 1.0)$ \\
        & & \textbf{MaF} & $\textbf{87.65}~(\pm 1.2)$& $74.25~(\pm 1.3)$& $76.13~(\pm 2.0)$& $74.92~(\pm 1.8)$& $70.47~(\pm 3.6)$& $74.27~(\pm 4.0)$& $73.59~(\pm 2.1)$\\
        & \multirow{2}{*}{\textbf{Allsides}} & \textbf{Acc} & $\textbf{87.26}~(\pm 0.2)$& $83.72~(\pm 4.0)$& $82.88~(\pm 5.1)$& $80.59~(\pm 6.3)$& $83.08~(\pm 4.0)$& $83.27~(\pm 4.2)$& $83.50~(\pm 3.5)$\\
        & & \textbf{MaF} & $\textbf{86.79}~(\pm 0.2)$& $83.10~(\pm 4.2)$& $81.86~(\pm 6.2)$& $78.98~(\pm 8.1)$& $82.39~(\pm 4.2)$& $82.28~(\pm 5.3)$& $82.64~(\pm 4.0)$\\
        \midrule[0.75pt]
        \multirow{4}{*}{\textbf{MD}} & \multirow{2}{*}{\textbf{SLN}} & \textbf{MiF} & $\textbf{94.22}~(\pm 1.2)$& $80.94~(\pm 5.5)$& $83.50~(\pm 5.7)$& $83.94~(\pm 4.7)$& $86.33~(\pm 2.1)$ & $82.67~(\pm 9.2)$& $79.89~(\pm 6.3)$\\
        & & \textbf{MaF} & $\textbf{94.18}~(\pm 1.1)$& $82.95~(\pm 4.4)$& $85.55~(\pm 4.4)$& $85.65~(\pm 3.4)$& $86.79~(\pm 1.9)$& $85.26~(\pm 6.2)$& $82.71~(\pm 4.1)$\\
        & \multirow{2}{*}{\textbf{LUN}} & \textbf{MiF} & $\textbf{71.28}~(\pm 1.7)$& $41.13~(\pm 5.8)$& $50.18~(\pm 6.3)$& $57.94~(\pm 4.1)$ & $48.78~(\pm 6.3)$& $53.52~(\pm 6.5)$& $63.27~(\pm 4.0)$\\
        & & \textbf{MaF} & $\textbf{69.82}~(\pm 1.2)$& $35.95~(\pm 7.3)$& $47.27~(\pm 7.3)$& $55.58~(\pm 4.6)$& $44.11~(\pm 9.0)$& $48.98~(\pm 7.9)$& $61.86~(\pm 4.4)$\\
        \midrule[0.75pt]
        \multirow{4}{*}{\textbf{RCVP}} & \multirow{2}{*}{\textbf{Random}} & \textbf{BAcc} & $\textbf{92.36}~(\pm 0.3)$& $91.29~(\pm 2.4)$& $91.35~(\pm 0.4)$& $91.34~(\pm 0.5)$ & $92.14~(\pm 0.5)$& $91.82~(\pm 0.8)$& $91.18~(\pm 1.5)$ \\
        & & \textbf{MaF} & $89.33~(\pm 0.4)$& $88.16~(\pm 2.5)$& $87.81~(\pm 0.8)$& $88.50~(\pm 0.4)$& $\textbf{89.35}~(\pm 0.7)$& $89.01~(\pm 1.0)$& $88.19~(\pm 1.6)$\\
        & \multirow{2}{*}{\textbf{Time-based}} & \textbf{BAcc} & $\textbf{94.46}~(\pm 0.4)$& $93.58~(\pm 1.4)$& $93.47~(\pm 0.5)$& $93.91~(\pm 0.5)$& $93.06~(\pm 1.7)$& $92.37~(\pm 2.2)$& $93.06~(\pm 1.0)$\\
        & & \textbf{MaF} & $\textbf{91.97}~(\pm 0.5)$& $90.60~(\pm 2.1)$& $90.73~(\pm 0.6)$& $91.29~(\pm 0.5)$& $90.06~(\pm 2.4)$& $88.56~(\pm 4.5)$& $90.21~(\pm 1.1)$\\
        \bottomrule[1.5pt]
    \end{tabular}
    \label{tab:ablation}
    }
\end{table*}

\subsection{Learning and Inference}
After a total of $\mathcal{P}$ KALM layers, we obtain the final document representation as $\Big[ \Tilde{\mathbf{t}}_L^{(\mathcal{P})}, \Tilde{\mathbf{g}}_L^{(\mathcal{P})}, \Tilde{\mathbf{k}}_L^{(\mathcal{P})} \Big]$. Given the document label $a \in \mathcal{A}$, the label probability is formulated as $p(a|\boldsymbol{d}) \propto \mathrm{exp}\big(\mathrm{MLP}_a([ \Tilde{\mathbf{t}}_L^{(\mathcal{P})}, \Tilde{\mathbf{g}}_L^{(\mathcal{P})}, \Tilde{\mathbf{k}}_L^{(\mathcal{P})} ])\big)$. We then optimize KALM with the cross entropy loss function. At inference time, the predicted label is $\mathrm{argmax}_a p(a|\boldsymbol{d})$.

\section{Experiment}

\subsection{Experiment Settings}

\paragraph{Tasks and Datasets.} We propose KALM, a general method for knowledge-aware long document understanding. We evaluate KALM on three tasks that especially benefit from external knowledge and broader context: political perspective detection, misinformation detection, and roll call vote prediction. We follow previous works to adopt SemEval \citep{DBLP:conf/semeval/KieselMSVACSP19} and Allsides \citep{DBLP:conf/acl/LiG19} for political perspective detection, LUN \citep{DBLP:conf/emnlp/RashkinCJVC17} and SLN \citep{rubin2016fake} for misinformation detection, and the 2 datasets proposed in \citet{DBLP:conf/acl/MouWCNHJH20} for roll call vote prediction. For external KGs, we follow existing works to adopt the KGs in KGAP \citep{DBLP:journals/corr/abs-2108-03861}, CompareNet \citep{DBLP:conf/acl/HuYZZTSD020}, and ConceptNet \citep{DBLP:conf/aaai/SpeerCH17} for the three tasks.

\paragraph{Baseline methods.} We compare KALM with three types of baseline methods for holistic evaluation: pretrained LMs, task-agnostic knowledge-aware methods, and task-specific models. For pretrained LMs, we evaluate RoBERTa \citep{DBLP:journals/corr/abs-1907-11692roberta}, Electra \citep{DBLP:conf/iclr/ClarkLLM20electra}, DeBERTa \citep{DBLP:conf/iclr/HeLGC21deberta}, BART \citep{DBLP:conf/acl/LewisLGGMLSZ20bart}, and LongFormer \citep{DBLP:journals/corr/abs-2004-05150longformer} on the three tasks. For task-agnostic baselines, we evaluate KGAP \citep{DBLP:journals/corr/abs-2108-03861}, GreaseLM \citep{GreaseLM}, and GreaseLM+ on the three tasks. Task-specific models are introduced in the following sections. For pretrained LMs, task-agnostic methods, and KALM, we run each method five times and report the average performance and standard deviation. For task-specific models, we compare with the results originally reported since we follow the exact same experiment settings and data splits.

\begin{figure*}[t]
    \centering
    \includegraphics[width=0.9\linewidth]{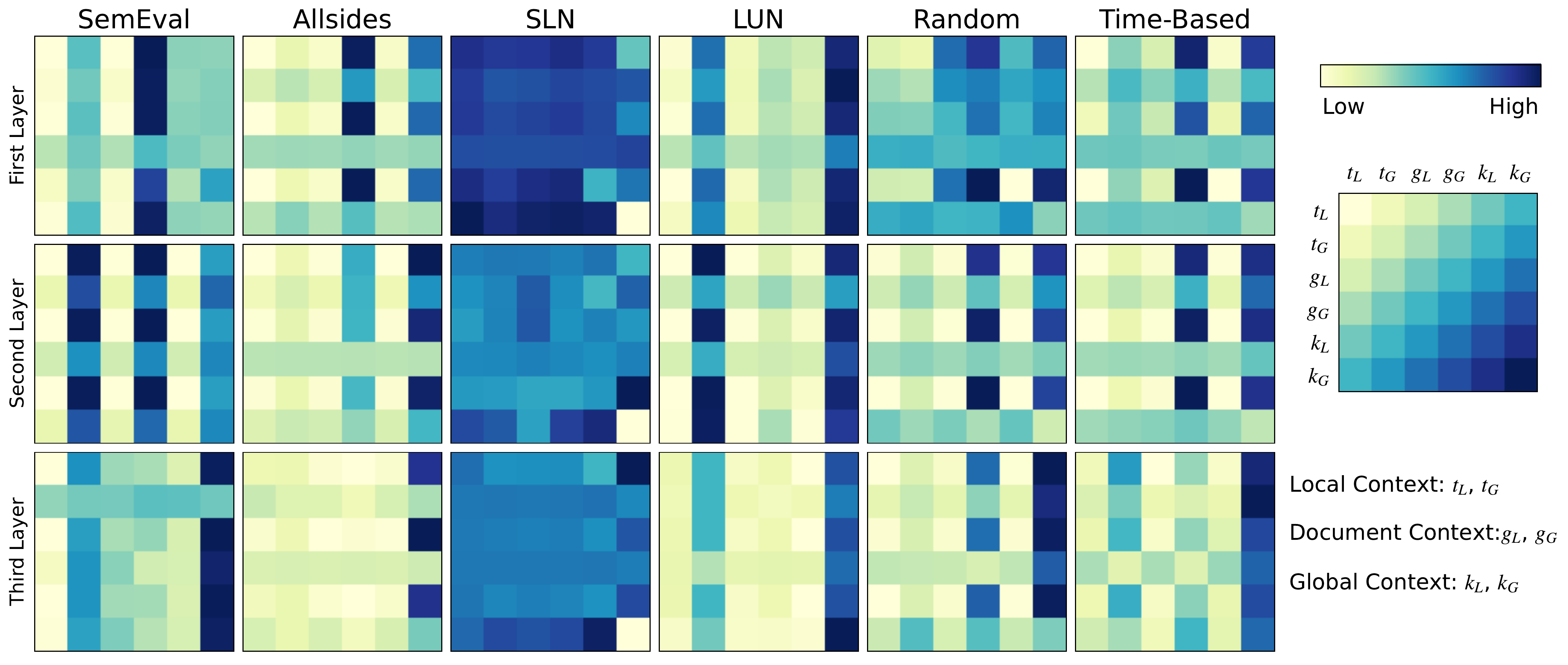}
    \caption{Interpreting the roles of contexts in the ContextFusion layer. $\mathbf{t}_L$, $\mathbf{t}_G$, $\mathbf{g}_L$, $\mathbf{g}_G$, $\mathbf{k}_L$, $\mathbf{k}_G$ denote the context representations in equations (9) and (10), so that the first two columns indicate how the local context attends to information in other contexts, the next two columns for the document context, and the last two for the global context.}
\label{fig:66}
\end{figure*}

\subsection{Model Performance}
We present the performance of task-specific methods, pretrained LMs, task-agnostic knowledge-aware baselines, and KALM in Table \ref{tab:big_table}. We select the best-performing task-specific baseline (Task SOTA) and pretrained language model (BestLM), while the full results are available in Tables \ref{tab:ppd}, \ref{tab:FND}, and \ref{tab:RCVP} in the appendix. Table \ref{tab:big_table} demonstrates that:

\begin{itemize}[leftmargin=*]
    \item KALM consistently outperforms all task-specific models, pretrained language models, and knowledge-aware methods on all three tasks and six datasets/settings. Statistical significance tests in Section \ref{subsec:significancetesting} further demonstrates KALM's superiority over existing models.
    \item Knowledge-aware LMs generally outperform pretrained LMs, which did not incorporate external knowledge bases in the pretraining process. This suggests that incorporating external knowledge bases could enrich document representations and boost downstream task performance.
    \item GreaseLM+ outperforms GreaseLM by adding the global context, which suggests the importance of jointly leveraging the three document contexts. KALM further introduces information exchange across contexts through the ContextFuion layer and achieves state-of-the-art performance. We further investigate the importance of three document contexts and the ContextFusion layer in Section \ref{subsec:contextfusion}.
\end{itemize}

\subsection{Context Exchange Study}
\label{subsec:exchange}
By jointly modeling three document contexts and employing the ContextFusion layer, KALM facilitates information exchange across the three document contexts. We conduct an ablation study to examine whether the contexts and the ContextFusion layer are essential in the KALM architecture. Specifically, we remove the three contexts one at a time and change the ContextFusion layer into MInt \citep{GreaseLM}, concatenation, and sum. Table \ref{tab:ablation} demonstrates that:
\begin{itemize}[leftmargin=*]
    \item All three levels of document contexts, local, document, and global, contribute to model performance. These results substantiate the necessity of jointly leveraging the three document contexts for long document understanding.
    \item When substituting our proposed ContextFusion layers with three existing combination strategies, MInt \citep{GreaseLM}, direct concatenation, and summation, performance drops are observed across multiple datasets. This suggests that the proposed ContextFusionn layer successfully boost model performance by enabling information exchange across contexts.
\end{itemize}


In addition to boosting model performance, the ContextFusion layer probes how different contexts contribute to document understanding. We calculate the average of attention weights' absolute values of the multi-head attention in the $\mathrm{TrmEnc}(\cdot)$ layer of ContextFusion and illustrate in Figure \ref{fig:66}, which shows that the three contexts' contribution and information exchange patterns vary with respect to datasets and KALM layers. Specifically, local and global contexts are important for the LUN dataset, document and global contexts are important for the task of roll call vote prediction, and the SLN dataset equally leverages the three contexts. However, for the task of political perspective detection, the importance of the three aspects varies with the depth of KALM layers. This is especially salient on SemEval, where KALM firstly takes a view of the whole document, then draws from both local and document-level contexts, and closes by leveraging global knowledge to derive an overall document representation. 

In summary, the ContextFusion layer in KALM successfully identifies the relative importance and information exchange patterns of the three contexts, providing insights into how KALM arrives at the conclusion and which context should be the focus of future research. We further demonstrate that the role and importance of each context change as training progresses in Section \ref{sec:contextexchangecont} in the appendix.

\begin{figure*}[t]
    \centering
    \includegraphics[width=1\linewidth]{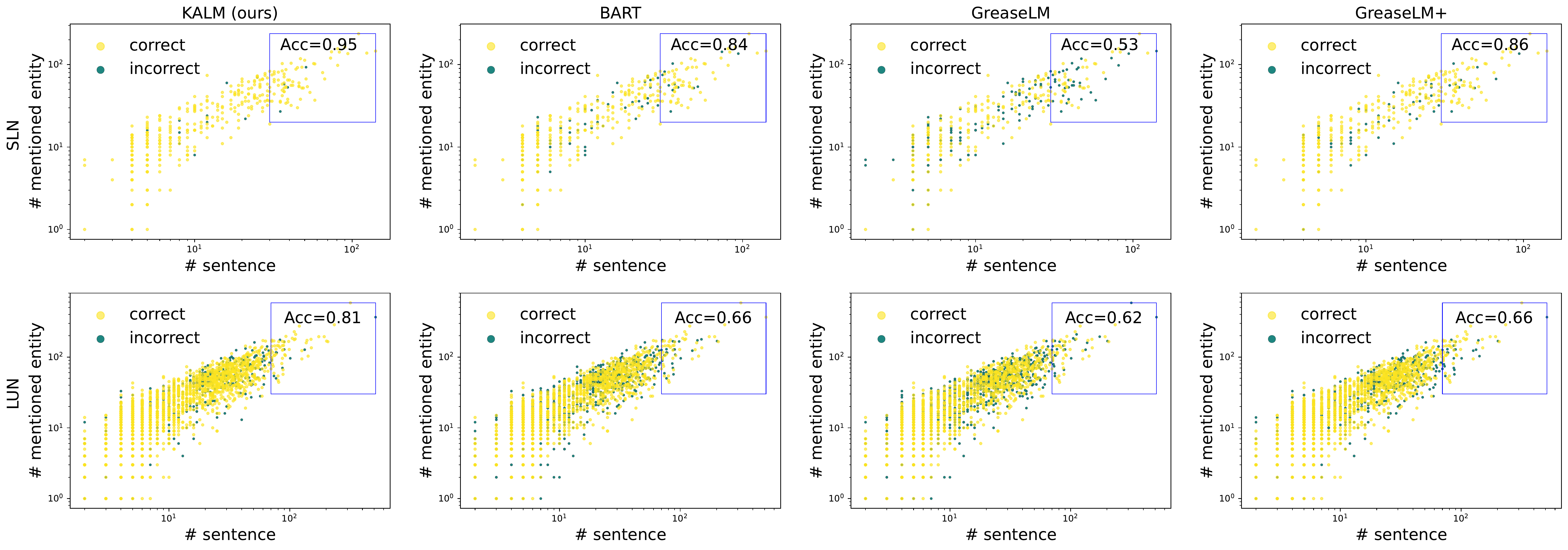}
    \caption{Error analysis of KALM and baseline methods. KALM successfully improves in the top-right corner, which represents documents with more sentences and more entailed knowledge.}
\label{fig:erroranalysis}
\end{figure*}

\subsection{Long Document Study}
KALM complements the scarce literature in knowledge-aware long document understanding. In addition to more input tokens, it often relies on more knowledge reference and knowledge reasoning. To examine whether KALM indeed improved in the face of longer documents and more external knowledge, we illustrate the performance of KALM and competitive baselines with respect to document length and knowledge intensity in Figure \ref{fig:erroranalysis}. Specifically, we use the number of mentioned entities to represent knowledge intensity and the number of sentences to represent document length, mapping each data point onto a two-dimensional space. It is illustrated that while baseline methods are prone to mistakes when the document is long and knowledge is rich, KALM alleviates this issue and performs better in the top-right corner. We further analyze KALM and more baseline methods' performance on long documents with great knowledge intensity in Figure \ref{fig:erroranalysiscont} in the appendix.

\subsection{Data Efficiency Study}
\label{subsec:dataefficiency}
\begin{figure*}[t]
    \centering
    \includegraphics[width=1\linewidth]{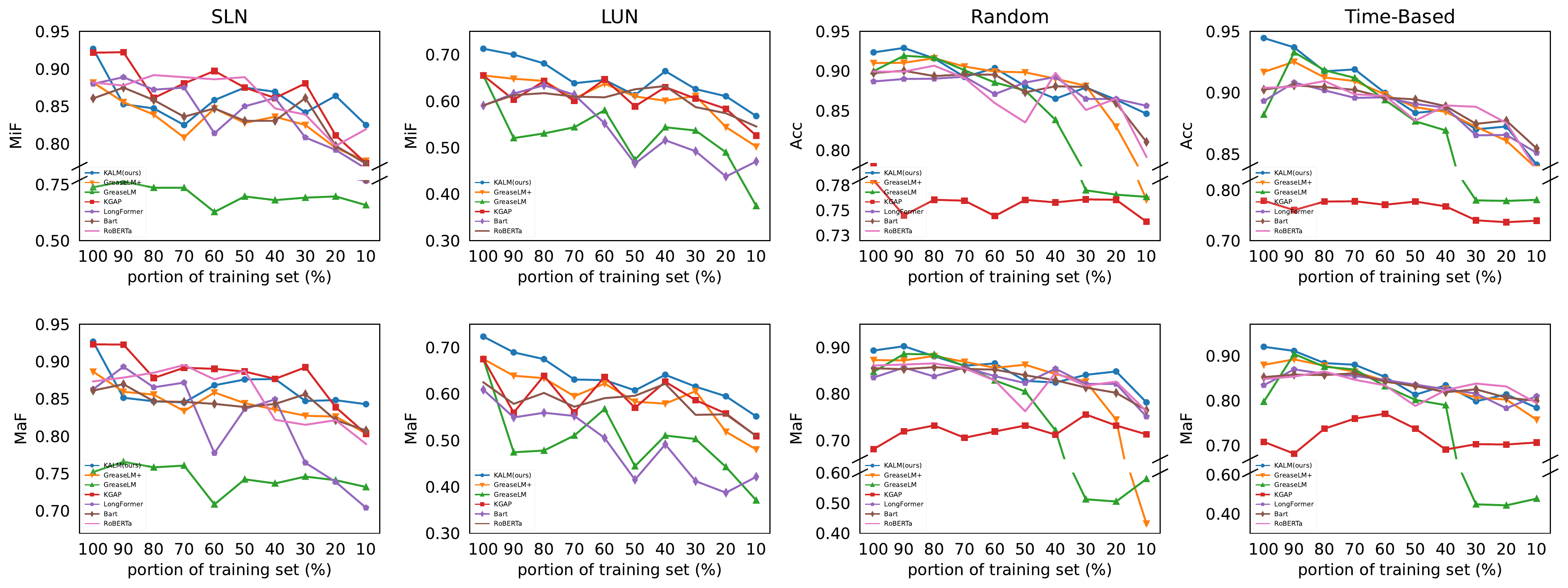}
    \caption{KALM and competitive baselines' performance when training data decreases from 100\% to 10\%. KALM maintains steady performance with as little as 10\% to 20\% of training data, while baseline methods witness serious performance deterioration.}
\label{fig:dataefficiency}
\end{figure*}

Existing works argue that introducing knowledge graphs to NLP tasks could improve data efficiency and help alleviate the need for extensive training data \citep{DBLP:conf/naacl/ZhangFCLLL22}. By introducing knowledge to all three document contexts and enabling knowledge-rich context information exchange, KALM might be in a better position to tackle this issue. To examine whether KALM has indeed improved data efficiency, we compare the performance of KALM with competitive baselines when trained on partial training sets and illustrate the results in Figure \ref{fig:dataefficiency}. It is demonstrated that while performance did not change greatly with 30\% to 100\% training data, baseline methods witness significant performance drops when only 10\% to 20\% of data are available. In contrast, KALM maintains steady performance with as little as 10\% of training data.

\section{Related Work}


Knowledge graphs are playing an increasingly important role in language models and NLP research. Commonsense \citep{DBLP:conf/aaai/SpeerCH17, DBLP:conf/esws/IlievskiSZ21, bosselut2019comet, west-etal-2022-symbolic, li-etal-2022-c3kg} and domain-specific KGs \citep{DBLP:journals/corr/abs-2108-03861, DBLP:journals/bioinformatics/LiZYDWWH22, gyori2017word} serve as external knowledge to augment pretrained LMs, which achieves state-of-the-art performance on question answering \citep{GreaseLM, DBLP:conf/naacl/YasunagaRBLL21, mitra2022constraint, bosselut2021dynamic, oguz-etal-2022-unik, feng-etal-2022-multi, heo2022hypergraph, ma-etal-2022-open, li2022dynamic, zhou2019multi}, social text analysis \citep{DBLP:conf/acl/HuYZZTSD020, DBLP:conf/naacl/ZhangFCLLL22, reddy2021newsclaims}, commonsense reasoning \citep{DBLP:conf/naacl/KimKKAHY22, DBLP:conf/acl/JungPCLKK022, amayuelas2021neural, liu2022generated}, and text generation \citep{rony-etal-2022-dialokg}. These approaches \citep{lu2022kelmother, zhang2019ernie, yu2022jaket, sun2020colake, yamada2020luke, qiu2019machine, xie2022unifiedskg} could be mainly categorized by the three levels of the context where knowledge injection happens.

\textbf{Local} context approaches focus on entity mentions and external knowledge in individual sentences to enable fine-grained knowledge inclusion. A straightforward way is to encode KG entities with KG embeddings \citep{DBLP:conf/nips/BordesUGWY13transe, DBLP:conf/aaai/LinLSLZ15transr, cucala2021explainable, DBLP:conf/iclr/SunDNT19} and infuse the embeddings with language representations \citep{DBLP:conf/acl/HuYZZTSD020, DBLP:journals/corr/abs-2108-03861, kang-etal-2022-kala}. Later approaches focus on augmenting pretrained LMs with KGs by introducing knowledge-aware training tasks and LM architectures \citep{Kepler, DBLP:conf/acl/WangTDWHJCJZ21, sridhar-yang-2022-explaining, moiseev-etal-2022-skill, kaur-etal-2022-lm, hu-etal-2022-knowledgeable, arora-etal-2022-metadata, de2021mention, meng2021mixture, he2021klmo}. Topic models were also introduced to enrich document representation learning \citep{gupta2018texttovectopic1, chaudhary2020explainabletopic2, wang2018topic3}. However, local context approaches fall short of leveraging inter-sentence and inter-entity knowledge, resulting in models that could not grasp the full picture of the text-knowledge interactions.

\textbf{Document-level} models analyze documents by jointly considering external knowledge across sentences and paragraphs. The predominant way of achieving document-level knowledge infusion is through "document graphs" \citep{DBLP:conf/naacl/ZhangFCLLL22}, where textual content, external knowledge bases, and other sources of information are encoded and represented as different components in graphs, often heterogeneous information networks \citep{DBLP:conf/acl/HuYZZTSD020, DBLP:journals/corr/abs-2108-03861, DBLP:conf/naacl/ZhangFCLLL22, yu-etal-2022-kg}. Graph neural networks are then employed to learn representations, which fuse both textual information and external KGs. However, document-level approaches fall short of preserving the original KG structure, resulting in models with reduced knowledge reasoning abilities.

\textbf{Global} context approaches focus on the KG, extracting relevant KG subgraphs based on entity mentions. Pruned with certain mechanisms \citep{DBLP:conf/naacl/YasunagaRBLL21} or not \citep{DBLP:conf/emnlp/QiuZFLJLLZ19}, these KG subgraphs are encoded with GNNs, and such representations are fused with LMs from simple concatenation \citep{DBLP:conf/acl/HuYZZTSD020} to deeper interactions \citep{GreaseLM}. However, global context approaches leverage external KGs in a stand-alone manner, falling short of enabling the dynamic integration of textual content and external KGs.

While existing approaches successfully introduced external KG to LMs, long document understanding poses new challenges to knowledge-aware NLP. Long documents possess greater knowledge intensity where more entities are mentioned, more relations are leveraged, and more reasoning is required to fully understand the nuances, while existing approaches are mostly designed for sparse knowledge scenarios. In addition, long documents also exhibit the phenomenon of knowledge co-reference, where central ideas and entities are reiterated throughout the document and co-exist in different levels of document contexts. In light of these challenges, we propose KALM to jointly leverage the local, document, and global contexts of long documents for knowledge incorporation.

\section{Conclusion}

In this paper, we propose KALM, a knowledge-aware long document understanding approach that introduces external knowledge to three levels of document contexts and enables interactive exchange across them. Extensive experiments demonstrate that KALM achieves state-of-the-art performance on three tasks across six datasets. Our analysis shows that KALM provides insights into the roles and patterns of individual contexts, improves the handling of long documents with greater knowledge intensity, and has better data efficiency than existing works.

\section*{Limitations}
Our proposed KALM has two limitations:

\begin{itemize}[leftmargin=*]
    \item KALM relies on existing knowledge graphs to facilitate knowledge-aware long document understanding. While knowledge graphs are effective and prevalent tools for modeling real-world symbolic knowledge, they are often sparse and hardly exhaustive \citep{DBLP:journals/corr/abs-2208-07622, pujara-etal-2017-sparsity}. In addition, external knowledge is not only limited to knowledge graphs but also exists in textual, visual, and other symbolic forms. We leave it to future work on how to jointly leverage multiple forms and sources of external knowledge in document understanding.
    \item KALM leverages TagMe \citep{ferragina2011fast} to identify entity mentions and build the three knowledge-aware contexts. While TagMe and other entity identification tools are effective, they are not 100\% correct, resulting in potentially omitted entities and external knowledge. In addition, running TagMe on hundreds of thousands of long documents is time-consuming and resource-consuming even if processed in parallel. We leave it to future work on how to leverage knowledge graphs for long document understanding without explicitly using entity linking tools.
\end{itemize}

\section*{Ethics Statement}
KALM is a knowledge-aware long document understanding approach that jointly leverages pretrained LMs and knowledge graphs on three levels of contexts. Consequently, KALM might exhibit many of the biases of the adopted language models \citep{liang2021towardslmbias1, nadeem2021stereosetlmbias2} and knowledge graphs \citep{Fisher2020kgbias1, Fisher2020kgbias2, mehrabi2021lawyerskgbias3, du-etal-2022-understanding, keidar-etal-2021-towards-automatic}. As a result, KALM might leverage the biased and unethical correlations in LMs and KGs to arrive at conclusions. We encourage KALM users to audit its output before using it beyond the standard benchmarks. We leave it to future work on how to leverage knowledge graphs in pretrained LMs with a focus on fairness and equity.

\section*{Acknowledgements}
We would like to thank the reviewers, the area chair, Vidhisha Balachandran, Melanie Sclar, and members of the Tsvetshop for their feedback. This material is funded by the DARPA Grant under Contract No. HR001120C0124. 
We also gratefully acknowledge support from NSF CAREER Grant No.~IIS2142739, the Alfred P.~Sloan Foundation Fellowship, and NSF grants No.~IIS2125201, IIS2203097, and IIS2040926.
Any opinions, findings and conclusions or recommendations expressed in this material are those of the authors and do not necessarily state or reflect those of the United States Government or any agency thereof.

\bibliography{custom}
\bibliographystyle{acl_natbib}

\appendix

\section{Additional Experiments}

\subsection{Context Exchange Study (cont.)}
\label{sec:contextexchangecont}
\begin{figure*}[t]
    \centering
    \includegraphics[width=0.9\linewidth]{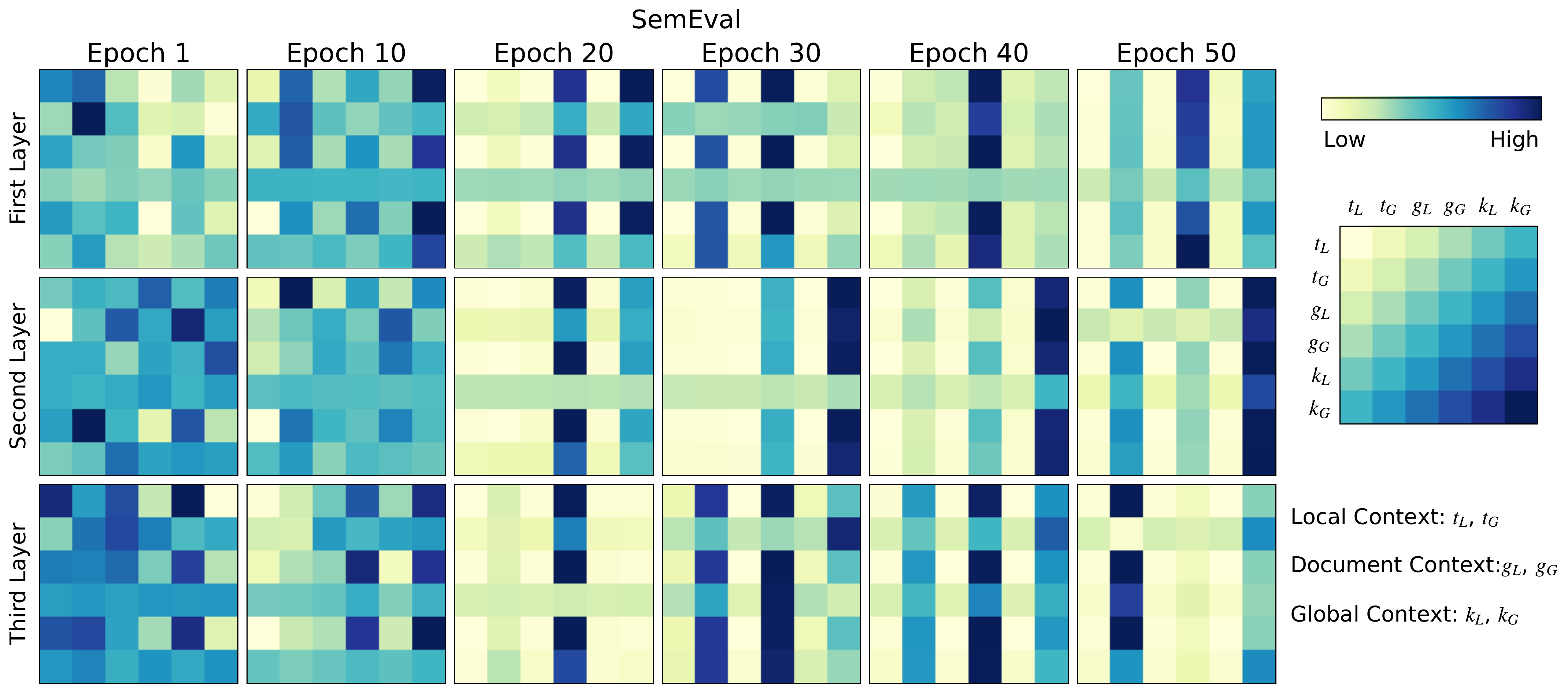}
    \caption{Interpreting the roles of the three contexts with respect to training progress on the SemEval dataset. $\mathbf{t}_L$, $\mathbf{t}_G$, $\mathbf{g}_L$, $\mathbf{g}_G$, $\mathbf{k}_L$, $\mathbf{k}_G$ denote the context representations in equations (9) and (10), so that the first two columns indicate how the local context attends to information in other contexts, the next two columns for the document context, and the last two columns for the global context.}
\label{fig:ex_epoch_sem}
\end{figure*}

In Section \ref{subsec:exchange}, we conducted an ablation study of the three knowledge-aware contexts and explored how the ContextFusion layer enables the interpretation of context contribution and information exchange patterns. It is demonstrated that the three contexts play different roles with respect to datasets and KALM layers. In addition, we explore whether the role and information exchange patterns of contexts change when the training progresses. Figure \ref{fig:ex_epoch_sem} illustrates the results with respect to training epochs, which shows that the attention matrices started out dense and ended sparse, indicating that the role of different contexts is gradually developed through time.

\subsection{Long Document Study (cont.)}
\begin{figure*}[t]
    \centering
    \includegraphics[width=\linewidth]{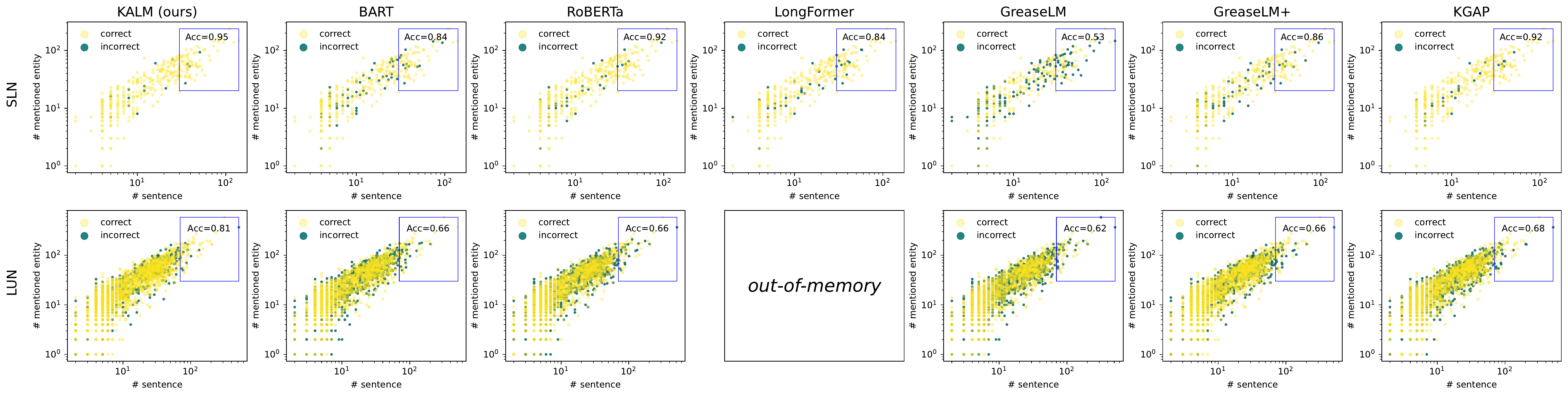}
    \caption{Error analysis of KALM and baselines. KALM successfully improves in the top-right corner, which represents documents with more sentences and more entailed knowledge.}
\label{fig:erroranalysiscont}
\end{figure*}

We present error analysis with respect to document length and knowledge intensity on more baseline methods, including language models (RoBERTa, BART, LongFormer), knowledge-aware LMs (KGAP, GreaseLM, GreaseLM+), and our proposed KALM in Figure \ref{fig:erroranalysiscont}. Our conclusion still holds true: KALM successfully improves performance on documents that are longer and contain more external knowledge, which are positioned in the top-right corner of the figure.

\subsection{Manual Error Analysis}
We manually examined 20 news articles from the LUN misinformation detection dataset where KALM made a mistake. Several news articles focused on the same topic of marijuana legalization, and some others focused on international affairs such as the conflict in Iraq. These articles feature entities and knowledge that are much more recent such as "pot-infused products" and "ISIS jihadists", which are emerging concepts and generally not covered by existing knowledge graphs. We present the relevant sentences in Table \ref{tab:manual_error_analysis}. This indicates the need for more comprehensive, up-to-date, and temporal knowledge graphs that grow with the world.

\renewcommand{\arraystretch}{1.1}
\begin{table*}[t]
    \centering
    \resizebox{\linewidth}{!}{
    \begin{tabular}{ll}
         \toprule[1.5pt] \textbf{Sample ID} & \textbf{Example Sentences}  \\ \midrule[0.75pt]
         \multirow{2}{*}{1853} & ... the legalization of recreational marijuana ... has created new markets for \textbf{pot-infused products} ...  \\
          & ... children who were taken to emergency departments due to accidental \textbf{THC} ingestion ... \\ \midrule[0.75pt]
         \multirow{2}{*}{1169} & Mr. Kerry met with Iraqi foreign minister \textbf{Hoshyar Zebari} about providing help in fighting the \textbf{ISIS jihadists} ... \\
          & ... territory north and north-east of Baghdad where the predominantly \textbf{Sunni militants} have penetrated within ... \\
          \bottomrule[1.5pt]
    \end{tabular}
    }
    \caption{Example sentences in the articles where KALM made a mistake. Emerging entities that are not covered by existing knowledge graphs are in \textbf{bold}.}
\label{tab:manual_error_analysis}
\end{table*}

\subsection{Significance Testing}
\label{subsec:significancetesting}
To examine whether KALM significantly outperforms baselines on the three tasks, we conduct one-way repeated measures ANOVA test for the results in Table \ref{tab:ppd}, Table \ref{tab:FND}, and Table \ref{tab:RCVP}. It is demonstrated that the performance gain is significant on five of the six datasets, specifically SemEval (against the second-best KCD on Acc and MaF), SLN (against the second-best KGAP on MiF and MaRecall), LUN (against the second-best CompareNet on MiF, MaF and MaRecall), Random (against the second-best GreasesLM+ on BAcc and MaF), and Time-Based (against the second-best GreaseLM+ on BAcc and MaF).

\subsection{Task-Specific Model Performance}
We present the full results for task-specific methods, pretrained language models, knowledge-aware task-agnostic models, and KALM on the three tasks and six datasets/settings in Tables \ref{tab:ppd}, \ref{tab:FND}, and \ref{tab:RCVP}.

\renewcommand{\arraystretch}{0.9}
\begin{table*}[t]
    \centering
    \caption{Model performance on the task of political perspective detection.}
    \resizebox{0.85\linewidth}{!}{
    \begin{tabular}{llcccc}
        \toprule[1.5pt]
        \multicolumn{2}{c}{\multirow{2}{*}{\textbf{Baseline}}} & 
        \multicolumn{2}{c}{\textbf{SemEval}} & \multicolumn{2}{c}{\textbf{Allsides}}\\
        \cmidrule(lr){3-4}
        \cmidrule(lr){5-6}
        \multicolumn{2}{c}{} & \textbf{Acc}& \textbf{MaF}& \textbf{Acc} & \textbf{MaF} \\ 
        \midrule[0.75pt]
        \multirow{3}{*}{\shortstack{\textbf{task}\\ \textbf{specific}}} & HLSTM & $81.71$ & $\slash$ & $76.45$& $74.95$\\ 
        & MAN & $86.21$& $84.33$& $85.00$& $84.25$\\ 
        & KCD &$89.90~(\pm 0.6)$ & $86.11~(\pm 1.1)$& $87.17~(\pm 0.2)$& $86.72~(\pm 0.3)$\\
        \midrule
        \multirow{5}{*}{\shortstack{\textbf{language}\\ \textbf{model}}} & RoBERTa& $85.56~(\pm 1.6)$& $77.94~(\pm 3.5)$& $68.71~(\pm 4.3)$& $65.39~(\pm 5.7)$\\ 
        & Electra& $78.87~(\pm 2.8)$& $62.85~(\pm 7.9)$& $63.14~(\pm 2.3)$& $58.24~(\pm 3.8)$\\ 
        & DeBERTa& $86.99~(\pm 1.9)$&  $80.62~(\pm 3.8)$& $67.86~(\pm 4.3)$& $63.50~(\pm 5.9)$\\ 
        & BART& $86.62~(\pm 1.5)$& $79.87~(\pm 2.6)$& $60.56~(\pm 3.8)$& $54.64~(\pm 5.4)$\\ 
        & LongFormer& $82.81~(\pm 2.3)$& $73.09~(\pm 4.5)$& $62.88~(\pm 3.0)$& $58.03~(\pm 4.6)$\\ 
        \midrule
        \multirow{9}{*}{\shortstack{\textbf{task}\\ \textbf{agnostic}}}
        & KELM&$86.40~(\pm 2.3)$ & $83.98~(\pm 1.0)$& $80.71~(\pm 2.4)$& $79.74~(\pm 2.7)$\\ 
        & KnowBERT-Wordnet&$81.71~(\pm 5.5)$ & $72.28~(\pm 6.7)$& $60.54~(\pm 0.4)$& $58.77~(\pm 0.6)$\\ 
        & KnowBERT-Wikidata&$76.72~(\pm 3.0)$ & $66.21~(\pm 5.0)$& $60.56~(\pm 0.7)$& $58.81~(\pm 0.5)$\\ 
        & KnowBERT-W+W&$84.73~(\pm 3.4)$ & $75.72~(\pm 5.3)$& $60.44~(\pm 0.3)$& $58.46~(\pm 0.5)$\\ 
        & Joshi et al.&$81.88~(\pm 2.1)$ & $77.15~(\pm 3.8)$& $80.88~(\pm 2.1)$& $79.73~(\pm 2.3)$\\ 
        & KGAP& $87.73~(\pm 1.8)$& $82.00~(\pm 3.1)$& $83.65~(\pm 1.3)$& $82.92~(\pm 1.4)$\\ 
        & GreaseLM& $86.64~(\pm 1.5)$& $80.32~(\pm 3.0)$& $80.23~(\pm 1.2)$& $79.17~(\pm 1.2)$\\
        & GreaseLM+&$85.66~(\pm 1.8)$ & $77.23~(\pm 4.1)$& $82.16~(\pm 5.5)$& $80.81~(\pm 7.1)$\\
        & KALM~(\textbf{Ours})& $\textbf{91.45}~(\pm 0.8)$& $\textbf{87.65}~(\pm 1.2)$& $\textbf{87.26}~(\pm 0.2)$& $\textbf{86.79}~(\pm 0.2)$\\
        \bottomrule[1.5pt]
    \end{tabular}
    \label{tab:ppd}
    }
\end{table*}

\begin{table*}[t]
    \centering
    \caption{Model performance on the task of misinformation detection.}
    \resizebox{\linewidth}{!}{
    \begin{tabular}{llcccccccc}
        \toprule[1.5pt]
        \multicolumn{2}{c}{\multirow{2}{*}{\textbf{Baseline}}} & 
        \multicolumn{4}{c}{\textbf{SLN}} & \multicolumn{4}{c}{\textbf{LUN}}\\ 
        \cmidrule(lr){3-6}
        \cmidrule(lr){7-10}
        \multicolumn{2}{c}{} & \textbf{MiF}& \textbf{MaPrecision} & \textbf{MaRecall} & \textbf{MaF} &\textbf{MiF}& \textbf{MaPrecision} & \textbf{MaRecall} & \textbf{MaF}\\ 
        \midrule[0.75pt]
        \multirow{5}{*}{\shortstack{\textbf{task}\\ \textbf{specific}}} & Rubin et al. & $\slash$ & $88.00$& $82.00$& $\slash$ & $\slash$ & $\slash$ & $\slash$ & $\slash$\\ 
        & Rashkin et al. & \slash& \slash& \slash& \slash& $\slash$ & $\slash$& $\slash$& $65.00$\\ 
        & GCN + Attn  & $85.27$& $85.59$& $85.27$& $85.24$& $67.08$& $68.60$& $67.00$& $66.42$\\
        & GAT + Attn  & $84.72$& $85.65$& $84.72$& $84.62$& $66.95$& $68.05$& $66.86$&$66.37$\\
        & CompareNet  & $89.17$& $89.82$& $89.17$& $89.12$& $69.05$& $72.94$& $69.04$&$68.26$\\
        \midrule[0.75pt]
        \multirow{5}{*}{\shortstack{\textbf{language}\\ \textbf{model}}} & RoBERTa& $88.17~(\pm 0.6)$& $89.02~(\pm 1.8)$& $88.17~(\pm 0.6)$& $87.34~(\pm 1.2)$& $59.09~(\pm 1.7)$& $62.49~(\pm 2.6)$& $59.11~(\pm 1.6)$& $55.52~(\pm 1.5)$\\ 
        & Electra& $75.44~(\pm 2.2)$& $83.22~(\pm 0.6)$& $75.44~(\pm 2.2)$& $67.53~(\pm 4.1)$& $60.10~(\pm 1.7)$& $63.26~(\pm 1.2)$& $60.11~(\pm 1.7)$& $58.57~(\pm 2.1)$\\ 
        & DeBERTa& $86.89~(\pm 6.6)$& $89.43~(\pm 3.7)$& $86.89~(\pm 6.6)$& $88.46~(\pm 4.9)$& $57.62~(\pm 3.1)$& $64.03~(\pm 0.9)$& $57.63~(\pm 3.1)$& $52.24~(\pm 5.3)$\\ 
        & BART& $86.06~(\pm 0.6)$& $86.13~(\pm 0.5)$& $86.06~(\pm 0.6)$& $86.12~(\pm 0.6)$& $59.05~(\pm 2.2)$& $60.89~(\pm 4.5)$& $59.07~(\pm 2.2)$& $54.18~(\pm 2.8)$\\ 
        & LongFormer& $88.00~(\pm 2.5)$& $88.84~(\pm 1.5)$& $87.44~(\pm 2.5)$& $86.29~(\pm 3.4)$& \multicolumn{4}{c}{out-of-memory}\\ 
        \midrule[0.75pt]
        \multirow{9}{*}{\shortstack{\textbf{task}\\ \textbf{agnostic}}}
        & KELM& $84.11~(\pm 0.6)$& $85.23~(\pm 0.7)$& $84.11~(\pm 0.6)$& $82.80~(\pm 1.3)$& $59.28~(\pm 2.1)$& $61.09~(\pm 2.8)$& $59.29~(\pm 2.1)$& $57.30~(\pm 1.6)$\\
        & KnowBERT-Wordnet& $74.72~(\pm 3.3)$& $77.22~(\pm 1.8)$& $74.72~(\pm 3.3)$& $72.74~(\pm 8.5)$& $55.63~(\pm 1.8)$& $56.29~(\pm 2.0)$& $55.63~(\pm 1.8)$& $55.02~(\pm 1.7)$\\
        & KnowBERT-Wikidata& $72.17~(\pm 2.5)$& $73.57~(\pm 0.6)$& $72.17~(\pm 2.5)$& $69.41~(\pm 6.9)$& $57.57~(\pm 0.5)$& $57.27~(\pm 0.6)$& $57.57~(\pm 0.5)$& $56.76~(\pm 0.6)$\\
        & KnowBERT-W+W& $78.67~(\pm 3.2)$& $79.36~(\pm 3.1)$& $78.67~(\pm 3.2)$& $79.80~(\pm 0.9)$& $65.52~(\pm 2.3)$& $67.50~(\pm 1.6)$& $65.53~(\pm 2.3)$& $63.94~(\pm 2.0)$\\
        & Joshi et al.& $92.72~(\pm 5.1)$& $84.95~(\pm 2.8)$& $83.37~(\pm 5.2)$& $83.98~(\pm 3.7)$& $58.57~(\pm 3.4)$& $62.56~(\pm 4.0)$& $58.59~(\pm 3.4)$& $56.73~(\pm 4.0)$\\
        & KGAP& $92.17~(\pm 1.2)$& $92.67~(\pm 0.9)$& $92.17~(\pm 1.2)$& $92.30~(\pm 0.9)$& $65.52~(\pm 2.3)$& $67.50~(\pm 1.6)$& $65.53~(\pm 2.3)$& $63.94~(\pm 2.9)$\\ 
        & GreaseLM&  $73.83~(\pm 0.9)$& $74.33~(\pm 0.8)$& $73.83~(\pm 0.9)$& $75.20~(\pm 0.8)$& $56.54~(\pm 1.5)$& $58.12~(\pm 2.7)$& $56.55~(\pm 1.5)$& $55.75~(\pm 1.6)$\\ 
        & GreaseLM+ &  $88.17~(\pm 0.8)$& $88.56~(\pm 0.6)$& $88.17~(\pm 0.8)$& $88.64~(\pm 0.6)$& $64.29~(\pm 2.4)$& $65.13~(\pm 2.7)$& $64.31~(\pm 2.4)$& $62.65~(\pm 3.7)$\\
        & KALM~(\textbf{Ours})& $\textbf{94.22}~(\pm 1.2)$& $\textbf{94.33}~(\pm 1.1)$& $\textbf{94.22}~(\pm 1.1)$& $\textbf{94.18}~(\pm 1.1)$& $\textbf{71.28}~(\pm 1.7)$& $\textbf{72.33}~(\pm 2.7)$& $\textbf{71.29}~(\pm 1.7)$& $\textbf{69.82}~(\pm 1.2)$\\
        \bottomrule[1.5pt]
    \end{tabular}
    \label{tab:FND}
    }
\end{table*}

\renewcommand{\arraystretch}{0.9}
\begin{table*}[t]
    \centering
    \caption{Model performance on the task of roll call vote prediction.}
    \resizebox{0.85\linewidth}{!}{
    \begin{tabular}{llcccc}
        \toprule[1.5pt]
        \multicolumn{2}{c}{\multirow{2}{*}{\textbf{Baseline}}} & \multicolumn{2}{c}{\textbf{Random}} & \multicolumn{2}{c}{\textbf{Time-Based}}\\ 
        \cmidrule[0.75pt](lr){3-4}
        \cmidrule[0.75pt](lr){5-6}
        \multicolumn{2}{c}{} & \textbf{BAcc}& \textbf{MaF}& \textbf{BAcc} & \textbf{MaF} \\ 
        \midrule[0.75pt]
        \multirow{4}{*}{\shortstack{\textbf{task}\\ \textbf{specific}}} & ideal-point & $86.46$ & $80.02$ & $\slash$& $\slash$\\ 
        & ideal-vector & $87.35$& $80.15$& $81.95$& $75.49$\\ 
        & Vote & $90.22$& $84.92$& $89.76$& $84.35$\\
        & PAR & $90.33$& $\slash$& $89.92$& $\slash$\\
        \midrule[0.75pt]
        \multirow{5}{*}{\shortstack{\textbf{language}\\ \textbf{model}}} & RoBERTa& $89.94~(\pm 0.2)$& $86.10~(\pm 0.7)$& $90.40~(\pm 0.8)$& $84.78~(\pm 2.2)$\\ 
        & Electra& $87.47~(\pm 0.3)$& $80.23~(\pm 0.7)$& $88.92~(\pm 0.4)$& $82.50~(\pm 1.7)$\\ 
        & DeBERTa& $86.98~(\pm 0.4)$& $80.07~(\pm 1.2)$& $88.59~(\pm 0.1)$& $81.38~(\pm 1.0)$\\ 
        & BART& $89.76~(\pm 0.5)$& $85.52~(\pm 0.6)$& $90.25~(\pm 0.6)$& $85.21~(\pm 2.1)$\\ 
        & LongFormer& $88.69~(\pm 0.4)$& $83.52~(\pm 1.2)$& $89.32~(\pm 1.4)$& $83.42~(\pm 3.8)$\\ 
        \midrule[0.75pt]
        \multirow{9}{*}{\shortstack{\textbf{task}\\ \textbf{agnostic}}}
        & KELM& $89.13~(\pm 1.1)$& $84.76~(\pm 2.0)$& $90.80~(\pm 0.2)$& $86.62~(\pm 0.4)$\\ 
        & KnowBERT-Wordnet& $86.72~(\pm 0.9)$& $79.33~(\pm 2.4)$& $86.92~(\pm 0.6)$& $78.90~(\pm 1.9)$\\ 
        & KnowBERT-Wikidata& $85.98~(\pm 0.8)$& $78.48~(\pm 1.0)$& $86.45~(\pm 0.5)$& $78.21~(\pm 0.7)$\\ 
        & KnowBERT-W+W& $85.75~(\pm 1.0)$& $78.70~(\pm 2.4)$& $87.07~(\pm 1.0)$& $78.42~(\pm 2.2)$\\ 
        & Joshi et al.& $91.43~(\pm 0.5)$& $89.64~(\pm 0.6)$& $92.63~(\pm 1.6)$& $89.31~(\pm 2.4)$\\ 
        & KGAP& $77.98~(\pm 0.5)$& $68.11~(\pm 6.0)$& $77.90~(\pm 0.6)$& $70.81~(\pm 4.6)$\\ 
        & GreaseLM& $89.99~(\pm 1.5)$& $84.72~(\pm 3.0)$& $88.21~(\pm 2.7)$& $79.73~(\pm 7.4)$\\ 
        & GreaseLM+& $91.01~(\pm 0.2)$& $87.29~(\pm 0.3)$& $91.69~(\pm 0.1)$& $87.95~(\pm 0.3)$\\ 
        & KALM~(\textbf{Ours})& $\textbf{92.36}~(\pm 0.3)$& $\textbf{89.33}~(\pm 0.4)$& $\textbf{94.46}~(\pm 0.4)$& $\textbf{91.97}~(\pm 0.5)$\\
        \bottomrule[1.5pt]
    \end{tabular}
    \label{tab:RCVP}
    }
\end{table*}

\subsection{Is local context enough?}
Though long document understanding requires attending to a long sequence of tokens, it is possible that sometimes only one or two sentences would give away the label of the document. We examine this by removing the document-level and global contexts in KALM, leaving only the local context to simulate this scenario. Comparing the local-only variant with the full KALM, there are 14.78\%, 10.53\%, 8.21\%, 4.85\%, 1.4\%, and 3.18\% performance drops across the six datasets in terms of macro-averaged F1-score. As a result, it is necessary to go beyond local context windows in long document understanding.

\section{Experiment Details}

\subsection{Dataset Details}

We present important dataset details in Table \ref{tab:dataset_details}. We follow the exact same dataset settings and splits in previous works \citep{DBLP:conf/naacl/ZhangFCLLL22, DBLP:conf/acl/HuYZZTSD020, Feng2021LegislatorRL} for fair comparison.

\renewcommand{\arraystretch}{1.1}
\begin{table*}[t]
    \centering
    \resizebox{\linewidth}{!}{
    \begin{tabular}{llccccc}
         \toprule[1.5pt] \textbf{Task} & \textbf{Dataset} & \textbf{\# Document} & \textbf{\# Class} & \textbf{Class Distribution} & \textbf{Document Length} & \textbf{Originally Proposed In} \\ \midrule[0.75pt]
         \multirow{2}{*}{\textbf{PPD}} & \textbf{SemEval} & 645 & 2 & 407 / 238 & 793.00 $\pm$ 736.93 & \citet{DBLP:conf/semeval/KieselMSVACSP19} \\
          & \textbf{Allsides} & 10,385 & 3 & 4,164 / 3,931 / 2,290 & 1316.81 $\pm$ 2978.71 & \citet{DBLP:conf/acl/LiG19} \\ \midrule[0.75pt]
         \multirow{2}{*}{\textbf{MD}} & \textbf{SLN} & 360 & 2 & 180 / 180 & \multirow{2}{*}{551.32 $\pm$ 661.82} & \citet{rubin2016fake} \\
          & \textbf{LUN} & 51,854 & 4 & 10,745 / 14,797 / 7,692 / 18,620 & & \citet{DBLP:conf/emnlp/RashkinCJVC17} \\ \midrule[0.75pt]
         \multirow{2}{*}{\textbf{RCVP}} & \textbf{random} & \multirow{2}{*}{1,155} & \multirow{2}{*}{2} & \multirow{2}{*}{304,655 / 95,464} & \multirow{2}{*}{653.94 $\pm$ 424.32} & \multirow{2}{*}{\citet{DBLP:conf/acl/MouWCNHJH20}} \\
          & \textbf{time-based} &  &  &  &  &  \\ \bottomrule[1.5pt]
    \end{tabular}
    }
    \caption{Dataset statistics. The number of long documents and class distribution does not add up for RCVP since multiple legislators vote on the same legislation.}
\label{tab:dataset_details}
\end{table*}

\subsection{Baseline Details}
We compare KALM with pretrained language models, task-specific baselines, and task-agnostic knowledge-aware methods to ensure a holistic evaluation. In the following, we provide a brief description of each of the baseline methods. We also highlight whether one approach leverages knowledge graphs and the three document contexts in Table \ref{tab:baselinecheck}.

\renewcommand{\arraystretch}{1.1}
\begin{table*}[t]
    \centering
    \resizebox{0.8\linewidth}{!}{
    \begin{tabular}{lcccccc}
         \toprule[1.5pt] 
         \multicolumn{1}{c}{\multirow{2}{*}{\textbf{Hyperparameter}}} & \multicolumn{2}{c}{\textbf{PPD}} & \multicolumn{2}{c}{\textbf{MD}} & \multicolumn{2}{c}{\textbf{RCVP}} \\
         \cmidrule[0.75pt](lr){2-3}
         \cmidrule[0.75pt](lr){4-5}
         \cmidrule[0.75pt](lr){6-7}
         \multicolumn{1}{c}{} & \textbf{SemEval} & \textbf{Allsides} & \textbf{SLN} & \textbf{LUN} & \textbf{random} & \textbf{time-based} \\ \midrule[0.75pt]
         max epochs & 50 & 25 & 3 & 5 & \multicolumn{2}{c}{100} \\ 
         optimizer & \multicolumn{6}{c}{RAdam \citep{liu2019variance}} \\ 
         seed LM & \multicolumn{6}{c}{BART \citep{DBLP:conf/acl/LewisLGGMLSZ20bart}} \\
         KB embedding & \multicolumn{6}{c}{TransE \citep{DBLP:conf/nips/BordesUGWY13transe}} \\
         dimension of hidden layers & \multicolumn{2}{c}{512} & \multicolumn{2}{c}{512} & \multicolumn{2}{c}{128} \\ 
         learning rate & \multicolumn{2}{c}{1e-3} & \multicolumn{2}{c}{1e-3} & \multicolumn{2}{c}{1e-4} \\ 
         weight decay & \multicolumn{2}{c}{1e-5} & \multicolumn{2}{c}{1e-5} & \multicolumn{2}{c}{1e-5}\\ 
         \# KALM layers & \multicolumn{2}{c}{2} & \multicolumn{2}{c}{2} & \multicolumn{2}{c}{2} \\
         \# attention heads & \multicolumn{2}{c}{8} & \multicolumn{2}{c}{8} & \multicolumn{2}{c}{8}\\ 
         dropout & \multicolumn{2}{c}{0.5} & \multicolumn{2}{c}{0.5} & \multicolumn{2}{c}{0.5}\\ 
         batch size & \multicolumn{2}{c}{16} & \multicolumn{2}{c}{16} & \multicolumn{2}{c}{4} \\
         \bottomrule[1.5pt]
    \end{tabular}
    }
    \caption{Hyperparameter settings of KALM.}
    \label{tab:hyperparameter}
\end{table*}

\begin{itemize}[leftmargin=*]
    \item \textbf{HLSTM} \citep{DBLP:conf/naacl/YangYDHSH16hlstm} is short for hierarchical long short-term memory networks. It was used in previous works \citep{DBLP:conf/acl/LiG19, DBLP:conf/acl/LiG21man} for political perspective detection.
    \item \textbf{MAN} \citep{DBLP:conf/acl/LiG21man} proposes to leverage social and linguistic information to design pretraining tasks and fine-tune on the task of political perspective detection.
    \item \textbf{KCD} \citep{DBLP:conf/naacl/ZhangFCLLL22} proposes to leverage multi-hop knowledge reasoning with knowledge walks and textual cues with document graphs for political perspective detection.
    \item \citet{rubin2016fake} proposes the SLN dataset and leverages satirical cues for misinformation detection.
    \item \citet{DBLP:conf/emnlp/RashkinCJVC17} proposes the LUN dataset and argues that misinformation detection should have more fine-grained labels than true or false.
    \item \textbf{GCN} \citep{DBLP:conf/iclr/KipfW17gcn} and \textbf{GAT} \citep{DBLP:conf/iclr/VelickovicCCRLB18} are leveraged along with the attention mechanism by \citet{DBLP:conf/acl/HuYZZTSD020} for misinformation detection on graphs.
    \item \textbf{CompareNet} \citep{DBLP:conf/acl/HuYZZTSD020} proposes to leverage knowledge graphs and compare the textual content to external knowledge for misinformation detection.
    \item \textbf{Ideal-point} \citep{DBLP:conf/icml/GerrishB11idealpoint} and \textbf{ideal-vector} \citep{DBLP:conf/emnlp/KraftJR16idealvector} propose to use 1d and 2d representations of political actors for roll call vote prediction.
    \item \textbf{Vote} \citep{DBLP:conf/acl/MouWCNHJH20} proposes to jointly leverage legislation text and the social network information for roll call vote prediction.
    \item \textbf{PAR} \citep{Feng2021LegislatorRL} proposes to learn legislator representations with social context and expert knowledge for roll call vote prediction.
    \item \textbf{RoBERTa} \citep{DBLP:journals/corr/abs-1907-11692roberta}, \textbf{Electra} \citep{DBLP:conf/iclr/ClarkLLM20electra}, \textbf{DeBERTa} \citep{DBLP:conf/iclr/HeLGC21deberta}, \textbf{BART} \citep{DBLP:conf/acl/LewisLGGMLSZ20bart}, and \textbf{LongFormer} \citep{DBLP:journals/corr/abs-2004-05150longformer} are pretrained language models. We use the pretrained weights $\textit{roberta-base}$, $\textit{electra-small-discriminator}$, $\textit{deberta-v3-base}$, $\textit{bart-base}$, and $\textit{longformer-base-4096}$ in Huggingface Transformers \citep{wolf2020transformers} to extract sentence representations, average across the whole document, and classify with softmax layers.
    \item \textbf{KELM} \citep{agarwal2021knowledgekelm} proposes to generate synthetic pretraining corpora based on structured knowledge bases. In this paper, we further pretrained the \textit{roberta-base} checkpoint on the KELM synthetic corpus and report performance on downstream tasks.
    \item \textbf{KnowBERT} \citep{peters2019knowledgeknowbert} is one of the first works to leverage external knowledge bases to enrich language representations. We used the three pretrained models, KnowBERT-Wordnet, KnowBERT-Wikidata, and KnowBERT-W+W for document representation extraction and report performance on downstream tasks.
    \item \citet{joshi2020contextualized} proposes to learn contextualized language representations by adding Wikipedia text to the input sequences and jointly learning text representations. This is similar to KALM's setting with only the local context, where Wikipedia descriptions of entities are concatenated to input texts.
    \item \textbf{KGAP} \citep{DBLP:journals/corr/abs-2108-03861} proposes to construct document graphs to jointly encode textual content and external knowledge. Gated relational graph convolutional networks are then adopted for document representation learning.
    \item \textbf{GreaseLM} \citep{GreaseLM} proposes to encode textual content with language model layers, encode knowledge graph subgraphs with graph neural networks and KG embeddings, and adopt MInt layers to fuse the two for question answering. In this paper, we implement GreaseLM by using MInt layers to fuse the local and global contexts.
    \item \textbf{GreaseLM+} is our extended version of GreaseLM, which adds the document-level context while keeping the original MInt layer instead of our proposed ContextFusion layer.
    \item \textbf{KALM} is our proposed approach for knowledge-aware long document understanding. It jointly infuses the local, document-level, and global contexts with external knowledge graphs and adopts ContextFusion layers to derive an overarching document representation.
    
\end{itemize}

\renewcommand{\arraystretch}{1.1}
\begin{table*}[t]
    \centering
    \caption{Checklist of whether baselines leverage knowledge graphs and the three document contexts.}
    \resizebox{0.8\linewidth}{!}{
    \begin{tabular}{llcccc}
        \toprule[1.5pt]
        \multicolumn{2}{c}{\textbf{Baseline}} & \textbf{Knowledge} & \textbf{Local} & \textbf{Document} & \textbf{Global}\\ 
        \midrule[0.75pt]
        \multirow{12}{*}{\shortstack{\textbf{task}\\ \textbf{specific}}} & HLSTM \citep{DBLP:conf/naacl/YangYDHSH16hlstm} & \xmark & \cmark & \cmark & \xmark \\ 
        & MAN \citep{DBLP:conf/acl/LiG21man} & \xmark & \cmark & \cmark & \xmark \\ 
        & KCD \citep{DBLP:conf/naacl/ZhangFCLLL22} & \cmark & \cmark & \cmark & \xmark \\
        & \citet{rubin2016fake} & \xmark & \cmark & \cmark & \xmark \\
        & \citet{DBLP:conf/emnlp/RashkinCJVC17} & \xmark & \cmark & \cmark & \xmark \\
        & GCN + Attn \citep{DBLP:conf/iclr/KipfW17gcn} & \cmark & \cmark & \cmark & \xmark \\
        & GAT + Attn \citep{DBLP:conf/iclr/VelickovicCCRLB18} & \cmark & \cmark & \cmark & \xmark \\
        & CompareNet \citep{DBLP:conf/acl/HuYZZTSD020} & \cmark & \cmark & \cmark & \xmark \\
        & ideal-point \citep{DBLP:conf/icml/GerrishB11idealpoint} & \xmark & \cmark & \xmark & \xmark \\
        & ideal-vector \citep{DBLP:conf/emnlp/KraftJR16idealvector} & \xmark & \cmark & \xmark & \xmark \\
        & Vote \citep{DBLP:conf/acl/MouWCNHJH20} & \xmark & \cmark & \cmark & \xmark \\
        & PAR \citep{Feng2021LegislatorRL} & \cmark & \cmark & \cmark & \xmark \\
        \midrule[0.75pt]
        \multirow{5}{*}{\shortstack{\textbf{language}\\ \textbf{model}}} & RoBERTa \citep{DBLP:journals/corr/abs-1907-11692roberta} & \xmark & \cmark & \xmark & \xmark \\ 
        & Electra \citep{DBLP:conf/iclr/ClarkLLM20electra}& \xmark & \cmark & \xmark & \xmark \\ 
        & DeBERTa \citep{DBLP:conf/iclr/HeLGC21deberta}& \xmark & \cmark & \xmark & \xmark \\ 
        & BART \citep{DBLP:conf/acl/LewisLGGMLSZ20bart}& \xmark & \cmark & \xmark & \xmark \\ 
        & LongFormer \citep{DBLP:journals/corr/abs-2004-05150longformer}& \xmark & \cmark & \cmark & \xmark \\ 
        \midrule[0.75pt]
        \multirow{4}{*}{\shortstack{\textbf{task}\\ \textbf{agnostic}}}
        & KELM \citep{agarwal2021knowledgekelm}& \cmark & \cmark & \xmark & \xmark \\
        & KnowBERT \citep{peters2019knowledgeknowbert}& \cmark & \cmark & \xmark & \xmark \\
        & \citet{joshi2020contextualized}& \cmark & \cmark & \xmark & \xmark \\
        & KGAP \citep{DBLP:journals/corr/abs-2108-03861}& \cmark & \xmark & \cmark & \xmark \\ 
        & GreaseLM \citep{GreaseLM}& \cmark & \cmark & \xmark & \cmark\\
        & GreaseLM+ (ours)& \cmark & \cmark & \cmark & \cmark \\ 
        & \textbf{KALM} (ours)& \cmark & \cmark & \cmark & \cmark \\
        \bottomrule[1.5pt]
    \end{tabular}
    }
    \label{tab:baselinecheck}
\end{table*}

\subsection{Evaluation Metrics Details}
We adopted these evaluation metrics throughout the paper: Acc (accuracy), MaF (macro-averaged F1-score), MiF (micro-averaged F1-score), MaPrecision (macro-averaged precision), MaRecall (macro-averaged recall), and BAcc (balanced accuracy). These metrics are chosen based on which metrics are used in previous works regarding the three tasks.

\subsection{Hyperparameter Details}
We present KALM's hyperparameter settings in Table \ref{tab:hyperparameter}. We conduct hyperparameter searches for different datasets and report the best setups.

\subsection{Where did the numbers come from?}
For task-specific baselines, we directly use the results reported in previous works \citep{DBLP:conf/naacl/ZhangFCLLL22, DBLP:conf/acl/HuYZZTSD020, Feng2021LegislatorRL} since we follow the same experiment settings and the comparison is thus fair. For pretrained LMs and task-agnostic baselines, we run each method \textbf{five times} with different random seeds and report the average performance as well as standard deviation. Figure \ref{fig:dataefficiency} is an exception, where we only run each method one time due to computing constraints.

\subsection{More experiment details}
We provide more details about the experiments that are worth further explaining.
\begin{itemize}[leftmargin=*]
    \item Table \ref{tab:RCVP}: We implement pretrained LMs and task-agnostic baselines for roll call vote prediction by using them to learn representations of legislation texts, concatenate them with the legislator representations learned with PAR \citep{Feng2021LegislatorRL}, and adopt softmax layers for classification.
    \item Table \ref{tab:ablation}: We remove each context by only applying ContextFusion layers to the other two context representations. We follow the implementation of MInt described in \citet{GreaseLM}. We implement concat and sum by using the concatenation and summation of the three context representations as the overall document representation.
    \item Figure \ref{fig:66}: The multi-head attention in the ContextFusion layer provides a 6 $\times$ 6 attention weight matrix indicating how information flowed across different contexts. The six rows (columns) stand for the local view of the local context, the global view of the local context, the local view of the document-level context, the global view of the document-level context, the local view of the global context, and the global view of the global context, which are described in detail in Section \ref{subsec:contextfusion}. The values in each square are the average of the absolute values of the attention weights across all data samples in the validation set.
\end{itemize}

\subsection{Computational Resources Details}
We used a GPU cluster with 16 NVIDIA A40 GPUs, 1,988G memory, and 104 CPU cores for the experiments. Running KALM with the best parameters takes approximately 1.5, 16, 3, 4, 1, and 1 hour(s) for the six datasets (SemEval, Allsides, SLN, LUN, random, time-based).

\subsection{Scientific Artifact Details}
KALM is built with the help of many existing scientific artifacts, including TagMe \citep{ferragina2011fast}, pytorch \citep{paszke2019pytorch}, pytorch lightning \citep{Falcon_PyTorch_Lightning_2019}, transformers \citep{wolf2020transformers}, pytorch geometric \citep{fey2019pytorchgeometric}, sklearn \citep{scikit-learn}, numpy \citep{harris2020arraynumpy}, nltk \citep{bird2009naturalnltk}, OpenKE \citep{han2018openke}, and the three adopted knowledge graphs \citep{DBLP:journals/corr/abs-2108-03861, DBLP:conf/acl/HuYZZTSD020, DBLP:conf/aaai/SpeerCH17}. We commit to make our code and data publicly available upon acceptance to facilitate reproduction and further research.

\end{document}